\documentclass[10pt,twocolumn,letterpaper]{article}

\usepackage[pagenumbers]{cvpr} % To force page numbers, e.g. for an arXiv version
\usepackage{makecell}
\usepackage[dvipsnames]{xcolor}

\definecolor{cvprblue}{rgb}{0.21,0.49,0.74}
\usepackage[pagebackref,breaklinks,colorlinks,citecolor=cvprblue]{hyperref}

\usepackage{amsmath}
\usepackage{algorithm}
\usepackage{algpseudocode}

\usepackage{xcolor}
\definecolor{r1_color}{RGB}{0,0,0}

\def\paperID{9407} % *** Enter the Paper ID here
\def\confName{CVPR}
\def\confYear{2024}

\title{SVGDreamer: Text Guided SVG Generation with Diffusion Model}

\author{%
  Ximing Xing$^{1}$,
  Haitao Zhou$^{1}$,
  Chuang Wang$^{1}$,
  Jing Zhang$^{1}$,
  Dong Xu$^{2}$,
  Qian Yu$^{1}$\thanks{Corresponding author} \\
  $^{1}$Beihang University \quad $^{2}$The University of Hong Kong \\
  \texttt{\{ximingxing, qianyu\}@buaa.edu.cn} \quad
  \texttt{dongxu@cs.hku.hk}
}

\begin{document}
\maketitle
\begin{abstract}
Text-guided scalable vector graphics (SVG) synthesis has broad applications in icon and sketch generation. However, existing text-to-SVG methods often suffer from limited editability, suboptimal visual quality, and low sample diversity. To address these challenges, we propose \textbf{SVGDreamer}, a novel framework for text-guided vector graphics synthesis. 
Our method introduces a \textbf{semantic-driven image vectorization (SIVE)} process, which decomposes the generation procedure into foreground objects and background elements, thereby improving structural controllability and editability. In particular, SIVE incorporates attention-based primitive control and an attention-mask loss to facilitate fine-grained manipulation of individual vector elements. 
To further improve generation quality and diversity, we propose \textbf{Vectorized Particle-based Score Distillation (VPSD)}, which models SVGs as distributions over control points and colors. Compared with existing text-to-SVG optimization methods, VPSD alleviates over-smoothed shapes, over-saturated colors, limited diversity, and slow convergence. Moreover, VPSD leverages a reward model to reweight vector particles, leading to better visual aesthetics and faster convergence. 
Extensive experiments demonstrate that SVGDreamer consistently outperforms existing baselines in editability, visual quality, and diversity. Project page: \href{https://ximinng.github.io/SVGDreamer-project/}{https://ximinng.github.io/SVGDreamer-project/}
\end{abstract}    
\section{Introduction}
\label{sec:intro}
Scalable Vector Graphics (SVG) represent visual content using geometric primitives such as B\'ezier curves, polygons, and lines. Compared with raster images, SVGs are resolution-independent, compact in storage, and, more importantly, inherently editable, allowing users to conveniently select, modify, and recombine individual elements. These properties make SVGs particularly attractive for visual design applications such as logos, posters, icons, and sketches.

Recently, there has been growing interest in automatic vector graphics generation. Most existing approaches build upon the differentiable rasterizer DiffVG~\cite{diffvg_Li_2020} and optimize vector primitives under external supervision~\cite{clipdraw_frans_2022,Styleclipdraw_schaldenbr_2022,clipclop_mirowski_2022,Clipasso_vinker_2022,CLIPascene_vinker_2023,CLIPFont_Song_2022,vectorfusion_jain_2023,diffsketcher_xing_2023}. Early methods typically rely on CLIP-based supervision, where the CLIP model~\cite{CLIP_radford_2021} guides the optimization of vector parameters through text--image alignment~\cite{clipdraw_frans_2022,Styleclipdraw_schaldenbr_2022,clipclop_mirowski_2022,CLIPFont_Song_2022,Clipasso_vinker_2022,CLIPascene_vinker_2023}. More recently, the remarkable progress of text-to-image (T2I) diffusion models~\cite{GLIDE_2022_nichol,ldm_Rombach_2022,DALLE2_2022_ramesh,imagen_2022_saharia,deepfloydif_stability_2023} has inspired text-to-vector generation. Representative methods such as VectorFusion~\cite{vectorfusion_jain_2023} and DiffSketcher~\cite{diffsketcher_xing_2023} exploit pretrained T2I models as powerful priors, either by using generated raster images as optimization targets or by distilling diffusion priors into vector graphics generation. In general, T2I-based methods produce higher-quality results than earlier CLIP-based approaches.

Despite this progress, existing T2I-based text-to-SVG methods still suffer from two major limitations. First, they provide limited editability. In conventional vector design workflows, visual elements are usually created and manipulated individually. By contrast, current optimization-based methods synthesize all primitives jointly without explicitly disentangling foreground objects from the background, which often entangles semantic components and makes object-level editing difficult. Second, the visual quality and diversity of the generated SVGs remain unsatisfactory. Existing methods commonly extend Score Distillation Sampling (SDS)~\cite{dreamfusion_poole_2023} to vector graphics generation~\cite{vectorfusion_jain_2023,diffsketcher_xing_2023}. However, SDS is known to suffer from over-smoothed geometry, over-saturated colors, limited diversity, and slow convergence. Moreover, as a mode-seeking optimization objective, SDS tends to optimize vector parameters toward an averaged solution, which can suppress fine-grained details and even miss objects specified in the text prompt.

To address these issues, we propose \textbf{SVGDreamer}, a new framework for text-guided vector graphics generation that improves editability, visual quality, and diversity. To enhance editability, we introduce a \textbf{S}emantic-driven \textbf{I}mage \textbf{VE}ctorization (SIVE) process, which decomposes synthesis into foreground objects and background elements. Specifically, SIVE employs an attention-based primitive control strategy to assign vector primitives to different semantic components, and uses cross-attention maps queried by text tokens to initialize control points for individual objects and background regions. We further design an attention-mask loss to optimize graphic elements hierarchically, enabling better separation, controllability, and object-level editing of the generated SVGs.

To improve visual quality and diversity, we further propose Vectorized Particle-based Score Distillation (VPSD) for vector graphics refinement. Instead of optimizing a single deterministic set of control points and colors, VPSD models SVGs as distributions over vector parameters and estimates these distributions with a LoRA~\cite{lora_hu_2022}-based network aligned with a pretrained diffusion model. This formulation alleviates the over-smoothing, over-saturation, limited diversity, and slow convergence issues of SDS-based vector generation. In addition, we incorporate ReFL~\cite{imagereward_xu_2023} to refine the estimation network with reward feedback, further improving the perceptual quality and aesthetic appeal of the synthesized SVGs.

Extensive experiments on diverse text prompts and vector styles demonstrate that SVGDreamer consistently outperforms existing baselines in editability, visual quality, and diversity. Moreover, our method shows strong generalization ability across different categories of vector graphics.

In summary, our contributions are as follows:
\begin{itemize}
\item We propose \textbf{SVGDreamer}, a novel framework for text-guided SVG generation that explicitly targets the key challenges of editability, visual quality, and diversity.
\item We introduce a semantic-driven image vectorization process with attention-based primitive control and an attention-mask loss, enabling semantic decomposition and fine-grained editing of individual vector elements.
\item We propose Vectorized Particle-based Score Distillation, a distribution-based optimization framework for SVG refinement that improves generation quality, diversity, and convergence efficiency; combined with reward-guided refinement, it further enhances the visual aesthetics of generated SVGs.
\item We conduct extensive experiments demonstrating that SVGDreamer achieves superior performance over strong baselines and generalizes well across diverse vector graphic styles and content categories.
\end{itemize}

%-----------------------------------Figure--------------------------------------
\begin{figure*}[t]
\centering
\includegraphics[width=0.95\linewidth]{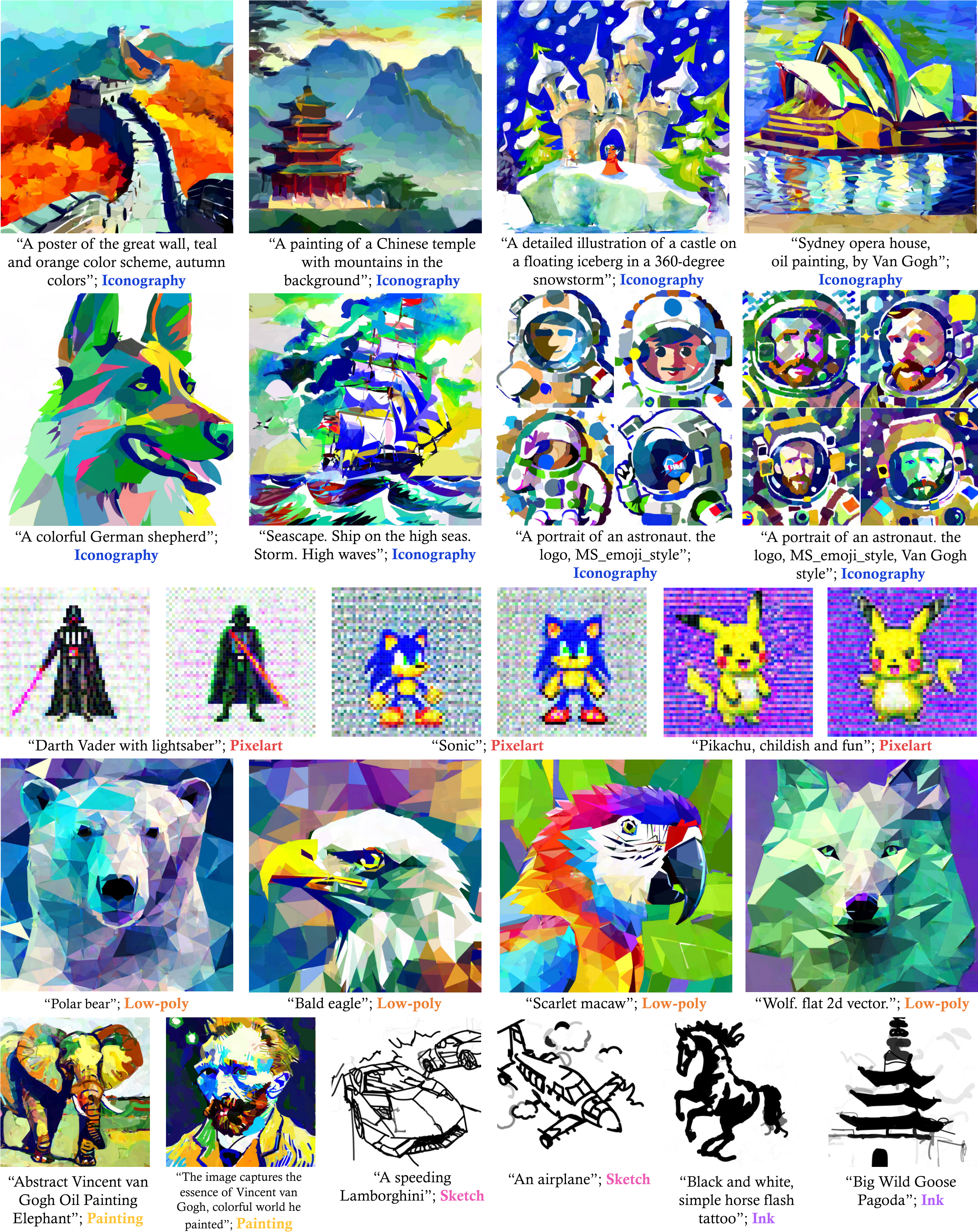}
\vspace{-0.5em}
\caption{
\textbf{SVGDreamer generates diverse SVGs from text prompts across multiple vector styles.} Different colored suffixes denote different styles. Our method flexibly controls style through vector primitives without relying on a fixed prompt suffix.
}
\label{fig:cases}
\end{figure*}
%-------------------------------------------------------------------------
\section{Related Work}
\label{sec:related_work}

\subsection{Vector Graphics Generation}

Scalable Vector Graphics (SVG) represent visual content using structured geometric primitives, such as B\'ezier curves, polygons, and lines. Early approaches to SVG generation mainly adopt sequence modeling frameworks to directly predict vector commands or token sequences~\cite{sketchrnn_David_2018,SVGVAE_Lopes_2019,deepsvg_carlier_2020,im2vec_reddy_2021,deepvecfont_wang_2021,textLogoLayout_wang_2022,iconshop_wu_2023}. Although these data-driven methods are effective in domains with abundant vector annotations, they typically depend on large-scale vector-form datasets and thus have limited generalization ability when synthesizing open-domain or structurally complex vector graphics.

An alternative line of work performs optimization-based vector graphics synthesis at inference time by matching rasterized outputs to target images or text guidance. DiffVG~\cite{diffvg_Li_2020} introduced a differentiable rasterizer that bridges vector graphics and raster images, making it possible to optimize vector primitives with image-space supervision. Based on this idea, a series of methods employ differentiable rendering for vector graphics synthesis and editing~\cite{ClipGen_Shen_2022,evolution_tian_2022,im2vec_reddy_2021,Styleclipdraw_schaldenbr_2022,LIVE_Ma_2022,marvel_su_2023,CLIPVG_song_2023,diffsketcher_xing_2023}. 

For text-guided vector generation, earlier methods often rely on CLIP~\cite{CLIP_radford_2021} as semantic supervision, leading to approaches such as CLIPDraw~\cite{clipdraw_frans_2022}, CLIP-CLOP~\cite{clipclop_mirowski_2022}, and CLIPasso~\cite{Clipasso_vinker_2022}. More recently, diffusion-based methods, such as VectorFusion~\cite{vectorfusion_jain_2023} and DiffSketcher~\cite{diffsketcher_xing_2023}, leverage pretrained text-to-image models to provide stronger visual priors for vector graphics generation, achieving promising results in applications such as icons, sketches, and pixel art. In contrast to these methods, our work focuses not only on generation quality but also on the editability and diversity of synthesized SVGs.

\subsection{Text-to-Image Diffusion Models}

Denoising diffusion probabilistic models (DDPMs)~\cite{diffusion_models_dickstein_2015,EestGrad_song_2019,ddpm_ho_2020,scorebased_song_2021} have achieved remarkable success in text-to-image synthesis. In particular, classifier-free guidance (CFG)~\cite{classifierfree_2022_ho} substantially improves generation fidelity and has become a standard component in modern text-conditioned diffusion frameworks, including GLIDE~\cite{GLIDE_2022_nichol}, Stable Diffusion~\cite{ldm_Rombach_2022}, DALL$\cdot$E 2~\cite{DALLE2_2022_ramesh}, Imagen~\cite{imagen_2022_saharia}, and DeepFloyd IF~\cite{deepfloydif_stability_2023}. 

Beyond raster image synthesis, pretrained text-to-image diffusion models have also enabled a broad range of text-guided generation tasks, such as text-to-3D~\cite{dreamfusion_poole_2023}. Their strong semantic alignment and visual prior make them particularly attractive for supervising vector graphics generation. Following this line of work, we use a pretrained Stable Diffusion model as the supervision source for text-to-SVG synthesis.

\subsection{Score Distillation Sampling}

Score Distillation Sampling (SDS), introduced in DreamFusion~\cite{dreamfusion_poole_2023}, provides an effective way to distill priors from pretrained 2D diffusion models into optimization-based generation tasks. It has been widely adopted in text-to-3D generation~\cite{dreamfusion_poole_2023,Magic3D_Lin_2023,SJC_Wang_2023}, where a parameterized representation is optimized using gradients derived from a text-conditioned diffusion model. Despite its effectiveness, SDS often suffers from over-smoothing, over-saturation, slow convergence, and limited diversity due to its deterministic and mode-seeking optimization behavior.

Inspired by SDS, recent text-to-SVG methods~\cite{vectorfusion_jain_2023,diffsketcher_xing_2023} also distill diffusion priors into vector graphics generation. However, they inherit similar limitations, often producing over-smoothed vector shapes and insufficiently diverse results. To mitigate these issues, Wang \textit{et al.}~\cite{prolificdreamer_wang_2023} proposed variational score distillation (VSD), which models the target representation as a random variable rather than a deterministic point estimate. Our proposed VPSD is closely related in spirit, but is specifically designed for SVG synthesis by modeling vector graphics as distributions over control points and colors, while further incorporating reward-guided refinement to improve visual quality and convergence.
\section{Methodology}
\label{sec:method}
In this section, we present \textbf{SVGDreamer}, an optimization-based framework for text-guided vector graphics synthesis. We represent an SVG as a set of paths $\{P_i\}_{i=1}^{n}$ and their corresponding color attributes $\{C_i\}_{i=1}^{n}$. Each path $P_i$ is parameterized by $m$ control points, i.e., $P_i=\{p_j\}_{j=1}^{m}=\{(x_j,y_j)\}_{j=1}^{m}$, and each path is associated with a color attribute $C_i=\{r,g,b,a\}_i$. The SVG parameters are denoted by $\theta=\{P_i,C_i\}_{i=1}^{n}$. Following prior work~\cite{diffvg_Li_2020}, we optimize $\theta$ by back-propagating gradients from rasterized images through a differentiable renderer $\mathcal{R}(\theta)$.

\begin{figure*}[t]
\centering
\includegraphics[width=1.0\linewidth]{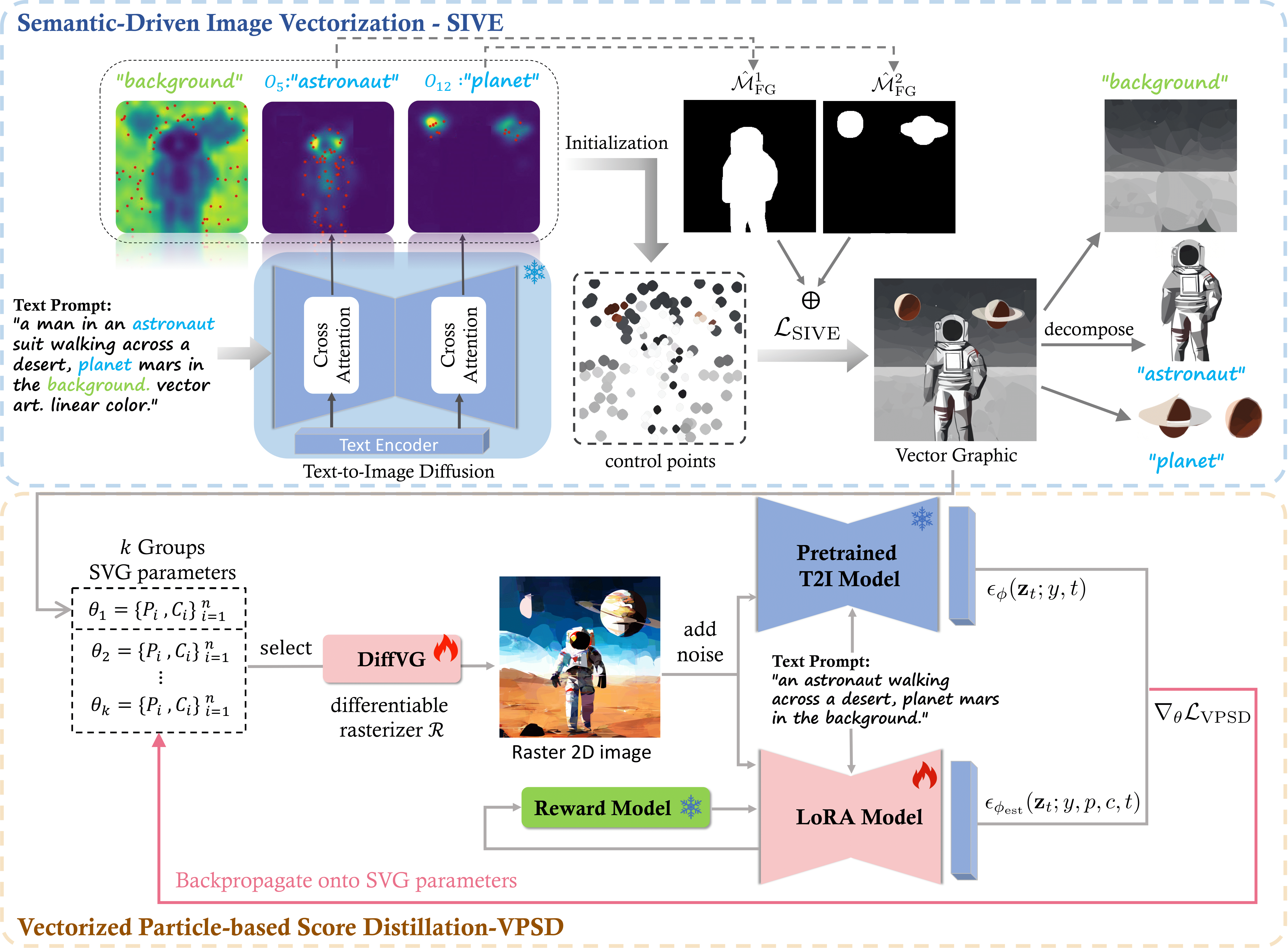}
\vspace{-1em}
\caption{
\textbf{Overview of SVGDreamer.} The method consists of two parts: semantic-driven image vectorization (SIVE, Sec.~\ref{sec:SIVE}) and SVG synthesis through VPSD optimization (Sec.~\ref{sec:SVGDreamer}). The result obtained from SIVE can be used as input of VPSD for further refinement.
} \label{fig:pipeline}
\vspace{-1em}
\end{figure*}

Our method leverages priors from a pretrained text-to-image diffusion model to guide the optimization of SVG parameters such that the rendered vector graphics match a given text prompt $y$. As shown in Fig.~\ref{fig:pipeline}, SVGDreamer consists of two key components: Semantic-driven Image Vectorization (SIVE) and Vectorized Particle-based Score Distillation (VPSD).

The first component, SIVE, aims to improve semantic decomposition and editability. It contains two stages: primitive initialization and semantic-aware optimization. Specifically, we extract cross-attention maps corresponding to different textual objects from the diffusion model, and use them to initialize control points and guide object-centric vectorization. This design enables the decomposition of the scene into foreground objects and background elements, thereby producing vector components that are more structurally separated and independently editable.

The second component, VPSD, is introduced to improve generation quality and diversity. Instead of optimizing a single deterministic set of vector parameters, VPSD models the distributions of path control points and color attributes to approximate the underlying SVG parameter distribution. This formulation enables the synthesis of diverse, high-quality vector graphics that better match the input text prompt.
%-------------------------------------Figure------------------------------------
%-------------------------------------------------------------------------
\subsection{SIVE: Semantic-driven Image Vectorization}
\label{sec:SIVE}
While rasterization is a well-studied operation in computer graphics, image vectorization---the inverse process of converting raster images into vector primitives---remains highly challenging. Given an input image, LIVE~\cite{LIVE_Ma_2022} progressively learns visual structures by recursively adding optimizable closed B\'ezier paths and jointly refining all paths. However, LIVE lacks explicit semantic decomposition and therefore struggles to distinguish different objects within a scene. As a result, similar or overlapping paths may be assigned to multiple visual subjects, leading to entangled vector representations. Moreover, LIVE-based methods~\cite{LIVE_Ma_2022,vectorfusion_jain_2023} are often insufficient for representing complex vector graphics composed of semantically distinct and structurally intricate elements.

To address these limitations, we propose semantic-driven image vectorization, which explicitly incorporates semantic cues into the vectorization process. SIVE consists of two stages: \emph{primitive initialization} and \emph{semantic-aware optimization}. In the first stage, we use attention maps to allocate distinct control points to image regions associated with different visual objects. In the second stage, we introduce an attention-mask loss to hierarchically optimize vector objects, improving their separation, controllability, and editability.

\subsubsection{Primitive Initialization}

Vectorizing multiple visual objects typically requires a large number of paths, which often leads to \emph{object-layer confusion} in LIVE-based methods, i.e., vector paths associated with different semantic objects become entangled during optimization. To alleviate this issue, we organize vector primitives according to semantic objects and initialize object-specific paths based on diffusion cross-attention maps.

Specifically, given a text prompt containing $O$ object tokens, we initialize $O$ groups of object-level control points according to the cross-attention maps of different objects. These maps are treated as foreground attention maps, denoted by $\mathcal{M}_{\mathrm{FG}}^i$, where $i$ indexes the $i$-th object token in the prompt. The remaining region is regarded as background. Accordingly, the foreground and background attention maps are defined as
\begin{equation}
\small
\mathcal{M}_{\mathrm{FG}}^i = \mathrm{softmax}\left(\frac{QK_i^{\top}}{\sqrt{d}}\right), \quad
\mathcal{M}_{\mathrm{BG}} = 1 - \sum_{i=1}^{O}\mathcal{M}_{\mathrm{FG}}^i,
\end{equation}
where $\mathcal{M}_{\mathrm{BG}}$ denotes the background attention map, $K_i$ is the key corresponding to the $i$-th text token, $Q$ denotes pixel-level query features, and $d$ is the latent projection dimension.

Following DiffSketcher~\cite{diffsketcher_xing_2023}, we normalize the attention maps and treat them as spatial sampling distributions to initialize vector primitives. Specifically, for each B\'ezier curve, we sample the first control point $p_{j=1}$ from the corresponding attention distribution. The remaining control points $\{p_j\}_{j=2}^{m}$ are then sampled within a small neighborhood around $p_{j=1}$ (with radius $0.05$ times the image size), yielding the initial object-aware path set.

\subsection{Vectorized Particle-based Score Distillation}
\label{sec:SVGDreamer}

Although rasterized diffusion samples provide useful guidance for SVG synthesis, directly vectorizing them is inherently lossy. Recent works~\cite{vectorfusion_jain_2023,diffsketcher_xing_2023} therefore adopt Score Distillation Sampling (SDS)~\cite{dreamfusion_poole_2023} to optimize SVG parameters with diffusion priors. Given a text prompt $y$, these methods directly optimize the SVG parameters $\theta=\{P_i,C_i\}_{i=1}^{n}$ through a differentiable renderer $\mathcal{R}(\theta)$. At each iteration, the renderer produces a rasterized image $\mathbf{x}=\mathcal{R}(\theta)$, which is further augmented to obtain $\mathbf{x}_a$. A pretrained latent diffusion model (LDM) $\epsilon_{\phi}$ then encodes $\mathbf{x}_a$ into the latent space using a VAE encoder~\cite{taming_esser_2021},
\begin{equation}
\mathbf{z}=\mathcal{E}(\mathbf{x}_a), \qquad \mathbf{z}\in\mathbb{R}^{(H/f)\times(W/f)\times 4},
\end{equation}
where $f$ is the encoder downsampling factor. Noise is then added to obtain
\begin{equation}
\mathbf{z}_t=\alpha_t \mathbf{z}+\sigma_t \boldsymbol{\epsilon}.
\end{equation}
The SDS gradient for updating $\theta$ is estimated as
\begin{equation}
\begin{split}
\nabla_{\theta} \mathcal{L}_{\mathrm{SDS}} & (\phi, \mathbf{x} = \mathcal{R}(\theta)) \triangleq 
\\
& \mathbb{E}_{t,\mathbf{\epsilon},a} 
\left[ 
w(t) (\mathbf{\epsilon}_{\phi} (\mathbf{z}_t;y,t) - \mathbf{\epsilon}) 
\frac{\partial \mathbf{z}}{\partial \mathbf{x}_a}
\frac{\partial \mathbf{x}_a}{\partial \theta}
\right]
\end{split}
\end{equation}
where $w(t)$ denotes the time-dependent weighting function.

Despite its effectiveness, SDS-based vector graphics generation often suffers from over-smoothed shapes, over-saturated colors, limited diversity, and slow convergence~\cite{dreamfusion_poole_2023,vectorfusion_jain_2023,diffsketcher_xing_2023,wordasimg_Iluz_2023}. These issues largely stem from the fact that SDS optimizes a single deterministic parameter set $\theta$ in a mode-seeking manner.

To address this limitation, we propose Vectorized Particle-based Score Distillation (VPSD), inspired by the variational score distillation framework~\cite{prolificdreamer_wang_2023}. Instead of representing an SVG as a single deterministic set of control points and colors, VPSD models SVG parameters as a conditional distribution $\mu(\theta|y)$ over possible vector graphics consistent with the text prompt $y$. Compared with SDS, which optimizes a single $\theta$, VPSD seeks to approximate the full distribution $\mu$, from which diverse SVG parameters can be sampled.

Following particle-based variational inference, we maintain $k$ groups of SVG parameters $\{\theta_i\}_{i=1}^{k}$ as particles to approximate $\mu(\theta|y)$. If optimization converges, these particles approximate samples from the target distribution $\mu^\ast$. VPSD relies on two score functions: a target score induced by the pretrained diffusion model, and a current score induced by the rendered SVG distribution. The target score of noisy real-image latents is given by the pretrained diffusion model $\epsilon_{\phi}(\mathbf{z}_t;y,t)$, while the current score is estimated by an auxiliary noise prediction network $\epsilon_{\phi_{\mathrm{est}}}(\mathbf{z}_t;y,p,c,t)$ trained on rendered images generated from the particle set $\{\theta_i\}_{i=1}^{k}$.

The VPSD gradient for updating the SVG particles is defined as
\begin{equation}
\begin{split}
& \nabla_{\theta} \mathcal{L}_{\mathrm{VPSD}} (\phi, \phi_\mathrm{est} , \mathbf{x} = \mathcal{R}(\theta)) \triangleq 
\\
& \mathbb{E}_{t,\epsilon,p,c} 
\left[ 
w(t) (
\mathbf{\epsilon}_{\phi} (\mathbf{z}_t;y,t) - 
\mathbf{\epsilon}_{\phi_\mathrm{est}}(\mathbf{z}_t;y,p,c,t) 
)
\frac{\partial \mathbf{z}}{\partial \theta}
\right]
\end{split}
\end{equation}
where $p$ and $c$ denote the path-control and color parameters associated with $\theta$, and $t\sim\mathcal{U}(0.05,0.95)$.

In practice, following~\cite{prolificdreamer_wang_2023}, we parameterize $\epsilon_{\phi_{\mathrm{est}}}$ as a LoRA~\cite{lora_hu_2022} adaptation of the pretrained diffusion model. The rendered samples produced by the current particle set are used to train the LoRA-based estimator with the denoising objective
\begin{equation}
\mathcal{L}_{\mathrm{LoRA}}
=
\mathbb{E}_{t,\boldsymbol{\epsilon},p,c}
\left\|
\epsilon_{\phi_{\mathrm{est}}}(\mathbf{z}_t;y,p,c,t)
-
\boldsymbol{\epsilon}
\right\|_2^2,
\end{equation}
where $\boldsymbol{\epsilon}$ denotes Gaussian noise. Only the LoRA parameters are updated, while the backbone parameters of the pretrained diffusion model remain frozen for computational efficiency.

Unlike~\cite{prolificdreamer_wang_2023}, which updates the estimator using randomly selected particles, we further introduce a \textbf{reward feedback learning} strategy to better exploit the learning status of SVG particles and accelerate convergence. As illustrated in Fig.~\ref{fig:vpsd_pipeline}, we employ a pretrained reward model~\cite{imagereward_xu_2023} to score samples generated by the LoRA-based estimator, and refine the estimator using reward-weighted feedback:
\begin{equation}
\mathcal{L}_{\mathrm{reward}}
=
\lambda
\mathbb{E}_{y}
\left[
\psi\big(
r(y,g_{\phi_{\mathrm{est}}}(y))
\big)
\right],
\end{equation}
where $g_{\phi_{\mathrm{est}}}(y)$ denotes the image generated by the estimator conditioned on prompt $y$, $r(\cdot,\cdot)$ is the pretrained reward model, $\psi(\cdot)$ is a reward-to-loss mapping function implemented with ReLU, and $\lambda=10^{-3}$. In the early stage of optimization, we use DDIM~\cite{ddim_song_2021} to efficiently generate $k$ samples. This reward-guided refinement improves the aesthetic quality of SVG synthesis and accelerates VPSD convergence.

Finally, VPSD updates the SVG particles using $\nabla_{\theta}\mathcal{L}_{\mathrm{VPSD}}$, while the estimator is optimized with
\begin{equation}
\mathcal{L}_{\mathrm{est}}
=
\mathcal{L}_{\mathrm{LoRA}}
+
\lambda_{\mathrm{r}}\mathcal{L}_{\mathrm{reward}},
\end{equation}
where $\lambda_{\mathrm{r}}$ controls the strength of reward feedback.

\subsection{Vector Representation Primitives}
\label{sec:various_primaries}

Beyond text prompts, SVGDreamer supports style control through different vector primitive configurations. Specifically, different visual styles are induced by constraining the primitive type and its parameterization. Users can therefore control the generated style either through textual descriptions or by selecting a suitable primitive set.

We consider six representative vector styles in this work:
\begin{itemize}
\item \textbf{Iconography.} A common SVG style composed of several filled paths with minimal yet expressive structures. We use closed B\'ezier curves with trainable control points and fill colors.
\item \textbf{Sketch.} A sparse and lightweight style for conveying shape with minimal visual elements. We use open B\'ezier curves with trainable control points and opacity values.
\item \textbf{Pixel Art.} A game-inspired style widely used for characters and backgrounds. We use square SVG polygons with fill colors.
\item \textbf{Low-Poly.} A style that approximates objects using a composition of simple geometric shapes. We use polygonal primitives with trainable vertex positions and fill colors.
\item \textbf{Painting.} A painterly vector style that mimics brush-based rendering. We use open B\'ezier curves with trainable control points, stroke colors, and stroke widths.
\item \textbf{Ink and Wash Painting.} A traditional Chinese art style characterized by varying black ink intensity. We use open B\'ezier curves with trainable control points, opacity values, and stroke widths.
\end{itemize}
\section{Experiments}
\label{sec:experiments}

%-----------------------------------Figure--------------------------------------
\begin{figure*}[h]
\centering
\includegraphics[width=1\linewidth]{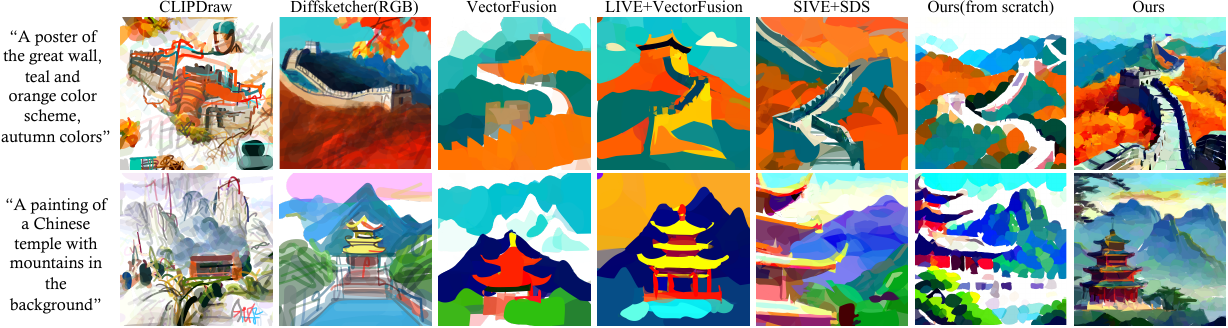}
\vspace{-1em}
\caption{
\textbf{Qualitative comparison with existing text-to-SVG methods.}
DiffSketcher was originally designed for vector sketch generation; therefore, we re-implemented it for RGB vector graphics generation.
}
\label{fig:qualitative_compare}
\vspace{-1em}
\end{figure*}
%-------------------------------------------------------------------------------
% --------------Table--------------
\begin{table*}[t]
\centering
\caption{Quantitative comparison of text-to-SVG generation methods.}
\vspace{-0.5em}
\resizebox{1.0\linewidth}{!}{
\begin{tabular}{l|cccccc}
\toprule
Method / Metric & FID~\cite{FID_Heusel_2017}$\downarrow$ & PSNR~\cite{PSNR_Hore_2010}$\uparrow$ & CLIPScore~\cite{CLIP_radford_2021}$\uparrow$ & BLIPScore~\cite{blip_li_2022}$\uparrow$ & Aesthetic~\cite{aesthetic_christoph_2022}$\uparrow$ & HPS~\cite{HPS_Wu_2023}$\uparrow$ \\
\midrule
CLIPDraw~\cite{clipdraw_frans_2022} & 160.64 & 8.35 & 0.2486 & 0.3933 & 3.9803 & 0.2347 \\
VectorFusion (scratch)~\cite{vectorfusion_jain_2023} & 119.55 & 6.33 & 0.2298 & 0.3803 & 4.5165 & 0.2334 \\
VectorFusion~\cite{vectorfusion_jain_2023} & 100.68 & 8.01 & 0.2720 & 0.4291 & 4.9845 & 0.2450 \\
DiffSketcher (RGB)~\cite{diffsketcher_xing_2023} & 118.70 & 6.75 & 0.2402 & 0.4185 & 4.1562 & 0.2423 \\
\midrule
\textbf{SVGDreamer} (from scratch) & 84.04 & 10.48 & 0.2951 & 0.4311 & 5.1822 & 0.2484 \\
\quad + Reward Feedback & 83.21 & 10.51 & 0.2988 & 0.4335 & 5.2825 & 0.2559 \\
\midrule
\textbf{SVGDreamer} & \textbf{59.13} & \textbf{14.54} & \textbf{0.3001} & \textbf{0.4623} & \textbf{5.5432} & \textbf{0.2685} \\
\bottomrule
\end{tabular}
} \label{tab:quantitative}
\vspace{-1em}
\end{table*}
% ----------------------------

\subsection{Qualitative Evaluation}

Figure~\ref{fig:qualitative_compare} compares SVGDreamer with existing text-to-SVG methods on representative prompts. Compared with CLIPDraw~\cite{clipdraw_frans_2022}, our method generates SVGs with substantially higher fidelity, clearer structure, and richer visual details. CLIPDraw often produces sparse and unstable path arrangements, which makes it difficult to form coherent vector objects under complex prompts. In contrast, SVGDreamer better preserves object boundaries and overall scene composition.

We further compare SVGDreamer with SDS-based methods~\cite{vectorfusion_jain_2023,diffsketcher_xing_2023}. These comparisons highlight the ability of our method to alleviate common SDS artifacts, including shape over-smoothing, ambiguous boundaries, and color over-saturation. In challenging scenes containing multiple semantic regions, baseline methods tend to produce averaged visual patterns and lose local structure. By contrast, our method preserves more recognizable object geometry and more stable color layouts.

As shown in the fifth column, SIVE improves semantic decomposition and editability by separating foreground objects from the background, but it does not fully overcome the intrinsic smoothness induced by SDS optimization. This observation verifies that semantic decomposition alone is not sufficient for high-quality SVG synthesis when the underlying optimization remains mode-seeking. In contrast, the final SVGDreamer results shown in the last two columns exhibit sharper structures, more faithful color composition, and better overall visual quality, regardless of whether optimization starts from scratch or from the full pipeline. These improvements also lead to higher aesthetic scores and more visually pleasing outputs.

%-----------------------------------Figure--------------------------------------
\begin{figure*}[t]
\centering
\includegraphics[width=1.0\linewidth]{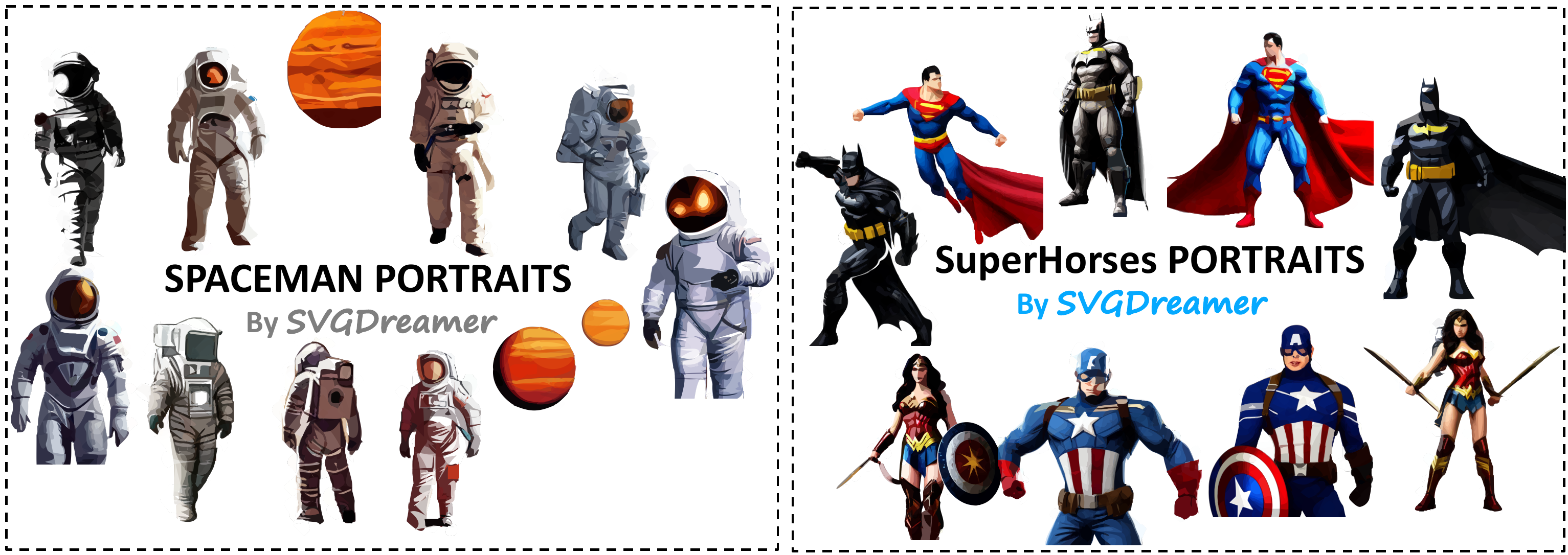}
\vspace{-1.5em}
\caption{
\textbf{Examples of editable vector assets generated by SVGDreamer.}
We specify the foreground content as an SVG asset through a text prompt. To generate assets that match a target SVG style (e.g., flat polygon vectors), we further constrain the vector representation with an appropriate prompt modifier, such as: \textit{... on a white background, full body action pose, complete body, concept art, flat 2d vector icon}.
}
\label{fig:editable}
\vspace{-0.5em}
\end{figure*}
%-------------------------------------------------------------------------

\subsection{Quantitative Evaluation}
\label{sec:qualitaitive}

We evaluate SVGDreamer from multiple perspectives, including Fr\'echet Inception Distance (FID)~\cite{FID_Heusel_2017}, Peak Signal-to-Noise Ratio (PSNR)~\cite{PSNR_Hore_2010}, CLIPScore~\cite{CLIP_radford_2021}, BLIPScore~\cite{blip_li_2022}, Aesthetic score~\cite{aesthetic_christoph_2022}, and Human Preference Score (HPS)~\cite{HPS_Wu_2023}. These metrics jointly measure generation diversity, reconstruction fidelity, text-image alignment, and perceptual quality. Table~\ref{tab:quantitative} compares our approach with representative text-to-SVG methods, including CLIPDraw~\cite{clipdraw_frans_2022}, VectorFusion~\cite{vectorfusion_jain_2023}, and DiffSketcher~\cite{diffsketcher_xing_2023}.

We conduct quantitative evaluation on the six vector styles introduced in~\cref{sec:various_primaries}. For each style, we use 10 text prompts and generate 50 SVGs per prompt, resulting in a diverse benchmark covering iconography, sketch, pixel art, low-poly, painting, and ink-wash painting settings. Such a protocol allows us to evaluate not only text-guided synthesis quality, but also the generalization ability of a method across substantially different primitive spaces and rendering styles.

To evaluate diversity and reconstruction fidelity, we use Stable Diffusion sampling results as reference images and report FID and PSNR, respectively. Although text-to-SVG generation is inherently open-ended, these reference images provide a consistent comparison target for all methods. The results in the first two columns show that SVGDreamer consistently outperforms competing methods in terms of both FID and PSNR. In particular, the significant gain in FID suggests that modeling SVGs as parameter distributions is more effective than deterministic SDS optimization for maintaining sample diversity. The improvement in PSNR further indicates that our method achieves more faithful reconstruction and more stable color composition than existing SDS-based approaches~\cite{vectorfusion_jain_2023,diffsketcher_xing_2023}.

To assess text-image alignment, we report both CLIPScore and BLIPScore. SVGDreamer achieves the best performance on both metrics, showing that the generated vector graphics are better aligned with the semantic content of the input text prompts. This advantage is especially important in multi-object scenes, where semantic entanglement in prior methods often weakens prompt fidelity.

To measure perceptual quality, we use the LAION aesthetic predictor~\cite{aesthetic_christoph_2022}. We additionally report HPS to evaluate preference from a human-centered aesthetic perspective. SVGDreamer again achieves the best scores. Notably, adding reward feedback further improves aesthetic score and HPS compared with the from-scratch version, which supports our claim that reward-guided particle reweighting helps the model converge toward more visually pleasing SVGs. Across all these metrics, SVGDreamer achieves the best overall performance, demonstrating its advantage in semantic alignment, visual quality, and diversity.

\subsection{Ablation Study}
\label{sec:abl_study}

\subsubsection{SIVE vs. LIVE~\cite{LIVE_Ma_2022}}
\label{sec:abl_live_sv_sive}

%-----------------------------------Figure--------------------------------------
\begin{figure}[h]
\centering
\includegraphics[width=1.0\linewidth]{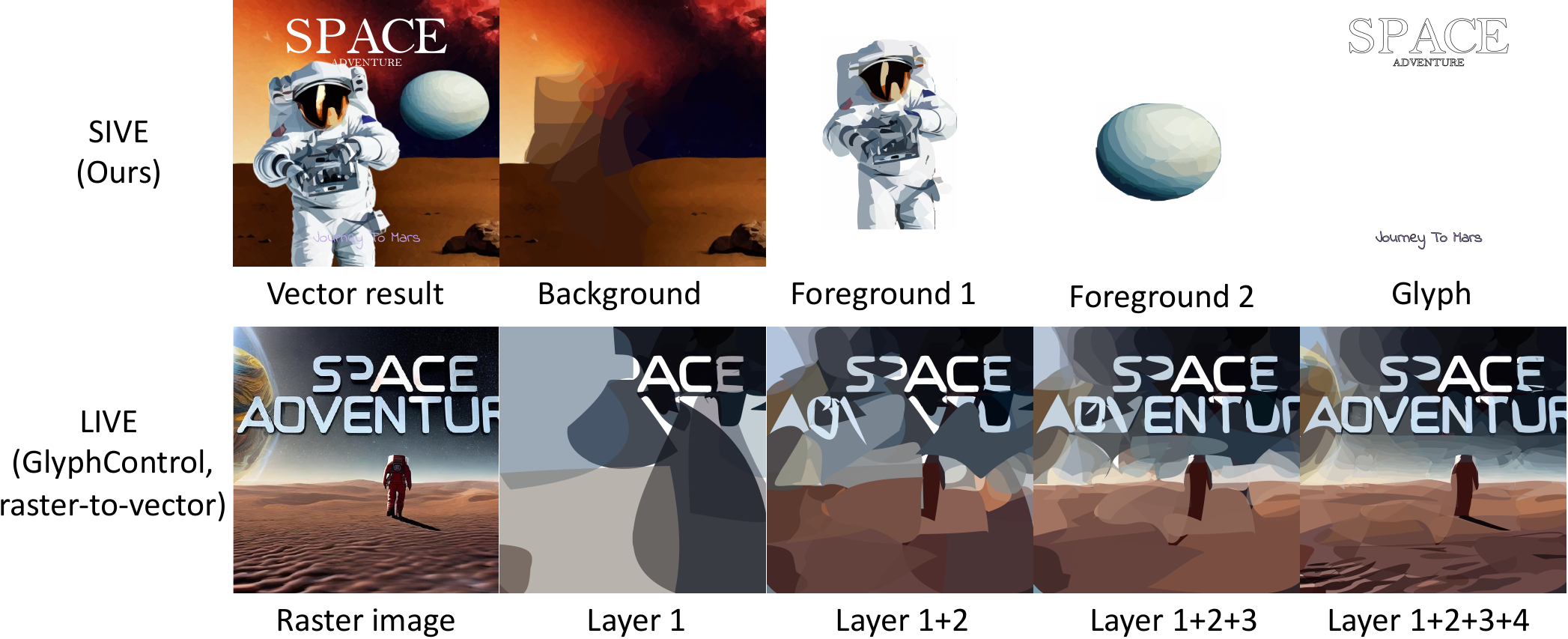}
\vspace{-1em}
\caption{
\textbf{Comparison between LIVE vectorization and SIVE.}
In the first row, ``Foreground 1'' and ``Foreground 2'' correspond to \textit{Astronaut} and \textit{Plants}, respectively. The glyphs are manually added and are not generated by our method. For LIVE, we follow the protocol of VectorFusion~\cite{vectorfusion_jain_2023}, which represents a vector image using 128 paths distributed over four layers, with 32 paths per layer.
}
\label{fig:abl_live_vs_sive}
\vspace{-1em}
\end{figure}
%-------------------------------------------------------------------------

LIVE~\cite{LIVE_Ma_2022} performs hierarchical, layer-wise vectorization by progressively optimizing vector paths. However, as shown in Fig.~\ref{fig:abl_live_vs_sive}, LIVE often fails to distinguish multiple semantic subjects within the same image, causing path hierarchies to overlap across different objects. This problem becomes more severe when representing complex vector graphics that require a large number of paths, resulting in redundant and entangled shapes that are difficult to edit. In practice, such entanglement not only reduces structural clarity, but also makes subsequent object-level manipulation cumbersome for human designers.

In contrast, SIVE produces more concise SVG structures with clearer semantic decomposition. By assigning paths to object-specific regions, SIVE enables object-level vectorization and improves the editability of the resulting SVGs. As illustrated in Fig.~\ref{fig:abl_live_vs_sive}, SIVE better preserves the semantic separation between foreground elements and background components, leading to cleaner path organization and more interpretable layer structure. This ablation validates that semantic-aware primitive initialization and optimization are essential for editable SVG synthesis.

\subsubsection{VPSD vs. LSDS~\cite{vectorfusion_jain_2023,wordasimg_Iluz_2023} and ASDS~\cite{diffsketcher_xing_2023}}

Text-to-SVG generation methods~\cite{vectorfusion_jain_2023,diffsketcher_xing_2023} are largely inspired by DreamFusion~\cite{dreamfusion_poole_2023}, but their SDS-based optimization often leads to over-smoothed vector shapes and limited diversity, similar to artifacts observed in early text-to-3D methods. The main difference between LSDS and ASDS lies in their augmentation strategies for the input image. Nevertheless, both remain fundamentally based on deterministic SDS optimization and are therefore still affected by mode-seeking behavior.

As shown in Table~\ref{tab:quantitative} and Fig.~\ref{fig:qualitative_compare}, VPSD consistently outperforms SDS-based variants in terms of FID, indicating stronger diversity and better resistance to mode collapse. VPSD also achieves higher PSNR, suggesting that it better preserves stable and visually coherent color composition without the over-saturation commonly observed in SDS-based methods. Qualitatively, VPSD produces SVGs with sharper local structures, more complete semantic objects, and fewer missing parts under complex prompts. These observations support the effectiveness of modeling vector graphics as parameter distributions instead of optimizing a single deterministic solution.

\subsection{Applications of SVGDreamer}

SVGDreamer generates highly editable vector graphics and can therefore serve as a practical tool for creating vector assets for poster and logo design. As shown in Fig.~\ref{fig:editable}, all graphical elements in the poster examples are generated by SVGDreamer. Designers can further recombine these elements with glyphs and layout components to create customized poster designs. Because SIVE explicitly separates foreground assets from the background and VPSD preserves stylistic consistency, the resulting SVG elements are more suitable for downstream composition than monolithic vector outputs.

In addition to poster design, the same pipeline can be used to create reusable SVG assets for icons, decorative elements, and concept-style illustrations. More application examples, including poster and logo design cases, are provided in the supplementary material.
\section{Conclusion}
\label{sec:conclusion}

In this paper, we presented \textbf{SVGDreamer}, a novel framework for text-guided vector graphics synthesis. SVGDreamer is built upon two key components: Semantic-Driven Image Vectorization (SIVE) and Vectorized Particle-Based Score Distillation (VPSD). Together, these designs enable the generation of SVGs with improved editability, higher visual quality, and greater diversity. We hope this work can advance text-to-SVG generation and broaden its applicability in real-world graphic design scenarios.

\noindent\textbf{Limitations.}
Our method still has several limitations. First, its editability is constrained by the decomposition capability of the underlying text-to-image (T2I) model. As a result, inaccurate or incomplete semantic separation in the T2I model may limit object-level controllability in the generated SVGs. We believe future advances in T2I diffusion models will further improve the semantic decomposition ability of our framework. Second, the number of control points for each object in SIVE is currently not determined adaptively. Developing an automatic mechanism for object-level primitive allocation is an important direction for future work.

\noindent\textbf{Acknowledgement.}\quad
This work was supported by the CCF-Baidu Open Fund Project and the Young Elite Scientists Sponsorship Program by CAST.

{
    \small
    \bibliographystyle{ieeenat_fullname}
    \bibliography{main}
}

\clearpage
\setcounter{page}{1}

\maketitlesupplementary

\appendix

\section*{Overview}
\label{sec:overview}

\begin{figure*}[h!]
\centering
\includegraphics[width=1.0\linewidth]{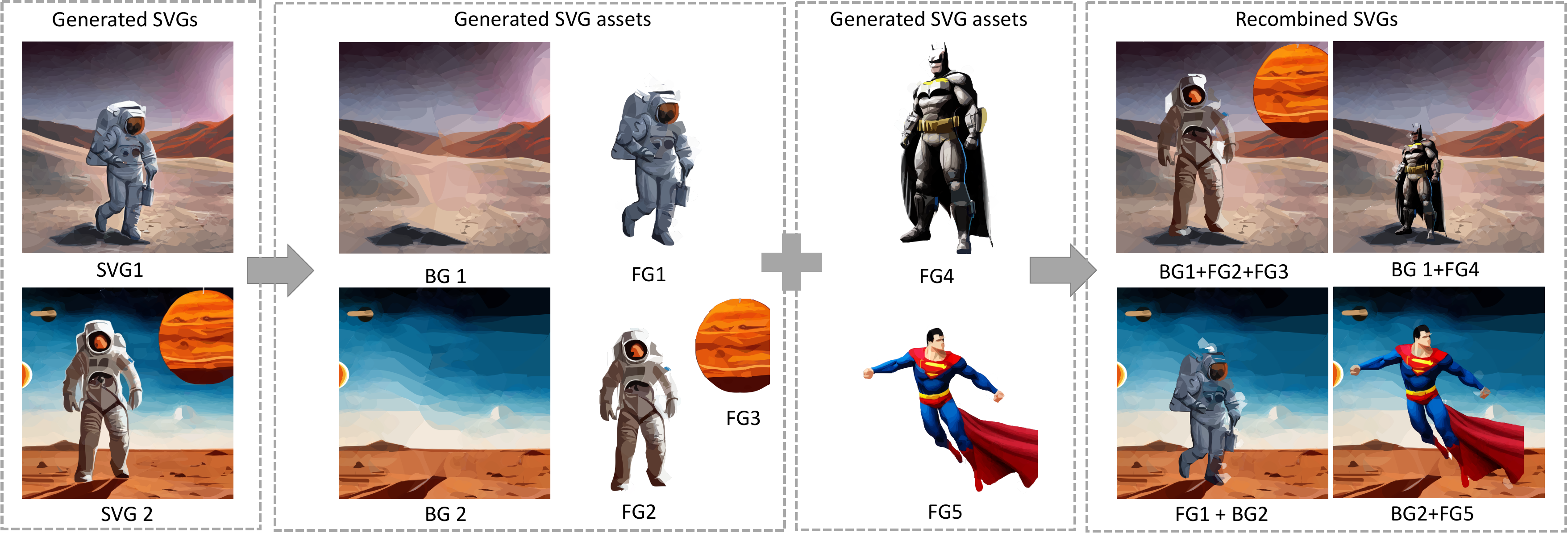}
\vspace{-1em}
\caption{
\textbf{Examples illustrating the editability of SVGs generated by SVGDreamer.}
The generated SVGs can be decomposed into reusable foreground and background assets, which can then be recombined across different samples to form new SVG compositions.
}
\label{fig:supp_editable}
\vspace{-1em}
\end{figure*}

This supplementary material provides additional results, analyses, and implementation details for SVGDreamer. Specifically, it contains the following:

\begin{itemize}
\item In Section~\ref{sec:supp_results}, we present more qualitative results of SVGDreamer, further demonstrating its editability, visual quality, and diversity.
\item In Section~\ref{sec:supp_application}, we showcase potential applications of SVGDreamer in poster and icon design.
\item In Section~\ref{sec:supp_implement}, we provide further implementation details.
\item In Section~\ref{sec:supp_prompt_sive}, we explain how semantic objects are identified in SIVE prompts.
\item In Section~\ref{sec:supp_ablation}, we present additional ablation studies on the effects of CFG weights (\cref{sec:supp_abl_CFG}), ReFL (\cref{sec:supp_abl_refl}), the number of vector particles (\cref{sec:supp_abl_n_particles}), and the number of paths (\cref{sec:supp_abl_num_paths}).
\item In Section~\ref{sec:supp_vpsd_2d_image}, we provide qualitative examples of applying VPSD to raster image synthesis.
\item In Section~\ref{sec:supp_algo_vpsd}, we present the pseudo-code of SVGDreamer. Our code is publicly available\footnote{\url{https://github.com/ximinng/SVGDreamer}}.
\end{itemize}

\section{Additional Qualitative Results}
\label{sec:supp_results}

\subsection{Editability.}

SVGDreamer is designed to generate editable vector graphics that can be reused beyond a single final output. As shown in Fig.~\ref{fig:editable} of the main paper and in the additional examples in Fig.~\ref{fig:supp_editable}, the generated SVGs can be decomposed into reusable foreground and background assets, and these assets can then be recombined across different samples to form new vector compositions.

Figure~\ref{fig:supp_editable} illustrates this process explicitly. Starting from two generated SVGs, SVGDreamer separates the background regions and the foreground objects into individual SVG assets. These separated elements remain semantically meaningful after decomposition: foreground assets such as characters preserve their identity, while background assets remain visually coherent as independent scene components. The decomposed assets can then be mixed across samples, producing new compositions such as placing a foreground object into a different background or combining multiple foreground elements within the same scene.

This property is particularly useful in practical design workflows, where designers often need to rearrange, replace, or reuse visual elements without redrawing them from scratch. Rather than treating generation as a one-shot process, SVGDreamer supports an asset-based workflow in which generated SVG components can be edited, recomposed, and integrated into downstream design tasks.

\subsection{Visual Quality and Diversity.}

\begin{figure*}[t]
\centering
\includegraphics[width=0.75\linewidth]{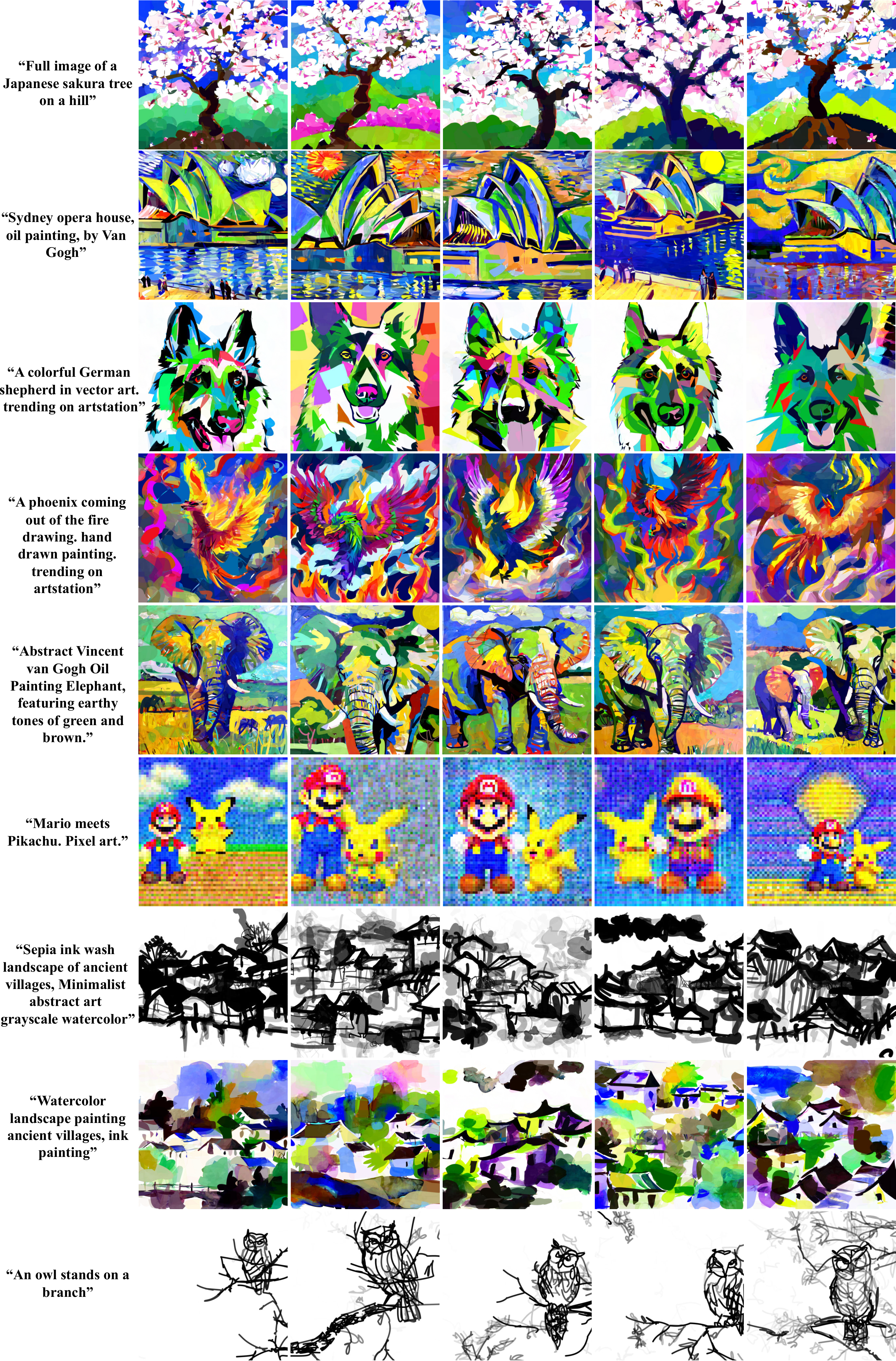}
\vspace{-0.5em}
\caption{
\textbf{Additional results generated by SVGDreamer.}
SVGDreamer supports both object-level and scene-level SVG synthesis across diverse styles. The visual style is controlled by the choice of vector primitives together with the text prompt.
} \label{fig:supp_more_vpsd_results}
\vspace{-1em}
\end{figure*}

Figure~\ref{fig:supp_more_vpsd_results} provides additional results generated by SVGDreamer across a wide range of prompts and primitive settings. The examples cover both object-centric generation (e.g., the German shepherd, phoenix, elephant, Mario and Pikachu, and owl) and scene-level generation (e.g., the cherry blossom tree, Sydney Opera House, ink-wash villages, and watercolor landscapes). These results show that SVGDreamer can preserve the main semantic structure of the prompt while adapting the rendering style to different vector primitive configurations.

A more specific observation from the figure is that different primitive settings lead to clearly different visual characteristics. Closed filled primitives produce compact color regions and are well suited for iconographic, oil-painting-like, and pixel-art results, while open curves and sparse strokes produce sketch-like or ink-wash appearances. For example, the ``Sydney Opera House'' and ``Abstract Vincent van Gogh oil painting elephant'' examples exhibit strong color blocking and stylized shape abstraction, whereas the ``owl'' and ``ink wash village'' examples are dominated by sparse line structures and monochrome brush-like marks. This indicates that the primitive design in SVGDreamer controls not only low-level appearance but also the overall style of the generated SVG.

The figure also shows that SVGDreamer supports style variation without requiring reference style images. Existing stylized vector generation methods, such as StyleCLIPDraw, typically rely on raster-style transfer pipelines and require an additional style reference. In contrast, our method specifies style directly through text prompts together with primitive constraints. For instance, prompts such as ``in Van Gogh's style,'' ``Pixel art,'' and ``ink wash'' lead to clearly different outputs while preserving the corresponding semantic content. This makes style control in SVGDreamer more direct and better suited to text-guided vector graphics synthesis.

\section{Applications of SVGDreamer}
\label{sec:supp_application}

In this section, we demonstrate how SVGDreamer can be applied to practical vector design tasks, including poster design and icon design. These examples highlight that SVGDreamer is not only a text-to-SVG generation framework, but also a useful tool for creating editable vector assets that can be further composed with typography, layout elements, and geometric templates in downstream design workflows.

\subsection{Poster Design.}

\begin{figure*}[t]
\centering
\includegraphics[width=0.95\linewidth]{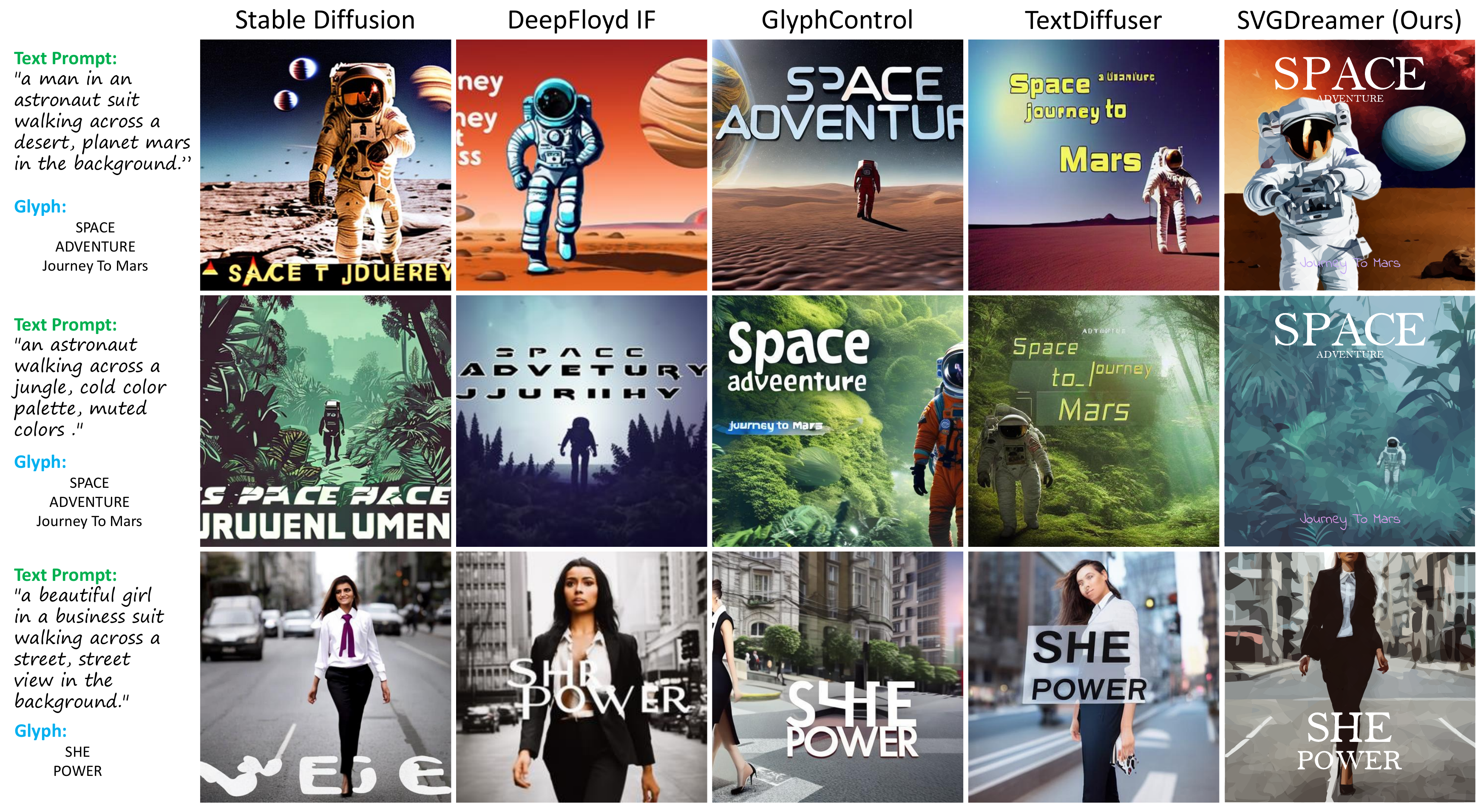}
\vspace{-0.5em}
\caption{
\textbf{Comparison of posters generated by different methods.}
The input prompts and glyphs are shown on the left. SVGDreamer generates clearer text and more editable poster layouts than raster text-to-image baselines.
}
\label{fig:supp_posters}
\vspace{-1em}
\end{figure*}

A poster typically combines typography and graphical content to convey information in a visually organized manner. Although text-to-image models have advanced rapidly, they still struggle with accurate text rendering, controllable typography, and stable layout composition. By contrast, SVG provides direct control over text placement, geometric transformation, and resolution-independent rendering. Figure~\ref{fig:supp_posters} compares posters created with SVGDreamer against those generated by four text-to-image baselines. Note that all outputs from the baseline methods are raster images.

Our poster design pipeline is asset-based. We first use SVGDreamer to generate the graphical content, and then use modern font libraries to create vectorized glyphs that can be precisely positioned, scaled, and transformed within the SVG canvas. Finally, we combine the generated vector graphics and vectorized text into a complete poster. Specifically, we use the FreeType font library\footnote{\url{http://freetype.org/index.html}} to represent glyphs as vector outlines composed of lines, B\'ezier curves, or B-spline curves. This representation allows text to be edited and rendered at arbitrary resolutions, making it naturally compatible with vector illustrations. Joint optimization of textual and graphical content remains an interesting direction for future work.

Figure~\ref{fig:supp_posters} shows that current text-to-image baselines still exhibit clear limitations for poster generation. Stable Diffusion~\cite{ldm_Rombach_2022} and DeepFloyd IF~\cite{deepfloydif_stability_2023} often generate severe text rendering errors, including missing letters, merged glyphs, repeated characters, and distorted word shapes. GlyphControl~\cite{glyphcontrol_yang_2023} improves controllability to some extent, but individual letters may still be omitted, and the inserted text can interfere with the underlying visual content. TextDiffuser~\cite{textdiffuser_chen_2023} supports more explicit layout control, yet the control masks often introduce visible artifacts and reduce the visual coherence between text and background imagery.

In contrast, SVGDreamer separates the generation of graphic content from the construction of typography and layout. This design avoids the text rendering failure modes of raster T2I systems and makes both components editable after generation. As a result, the posters generated by SVGDreamer exhibit clearer typography, more stable composition, and substantially greater flexibility for iterative design. This makes our approach more suitable for practical poster creation, where text, layout, and graphic elements often need to be refined repeatedly.

%-----------------------------------Figure--------------------------------------
%-------------------------------------------------------------------------
%-----------------------------------Figure--------------------------------------
%-------------------------------------------------------------------------

\subsection{Icon Design.}

\begin{figure}[h!]
\centering
\includegraphics[width=0.9\linewidth]{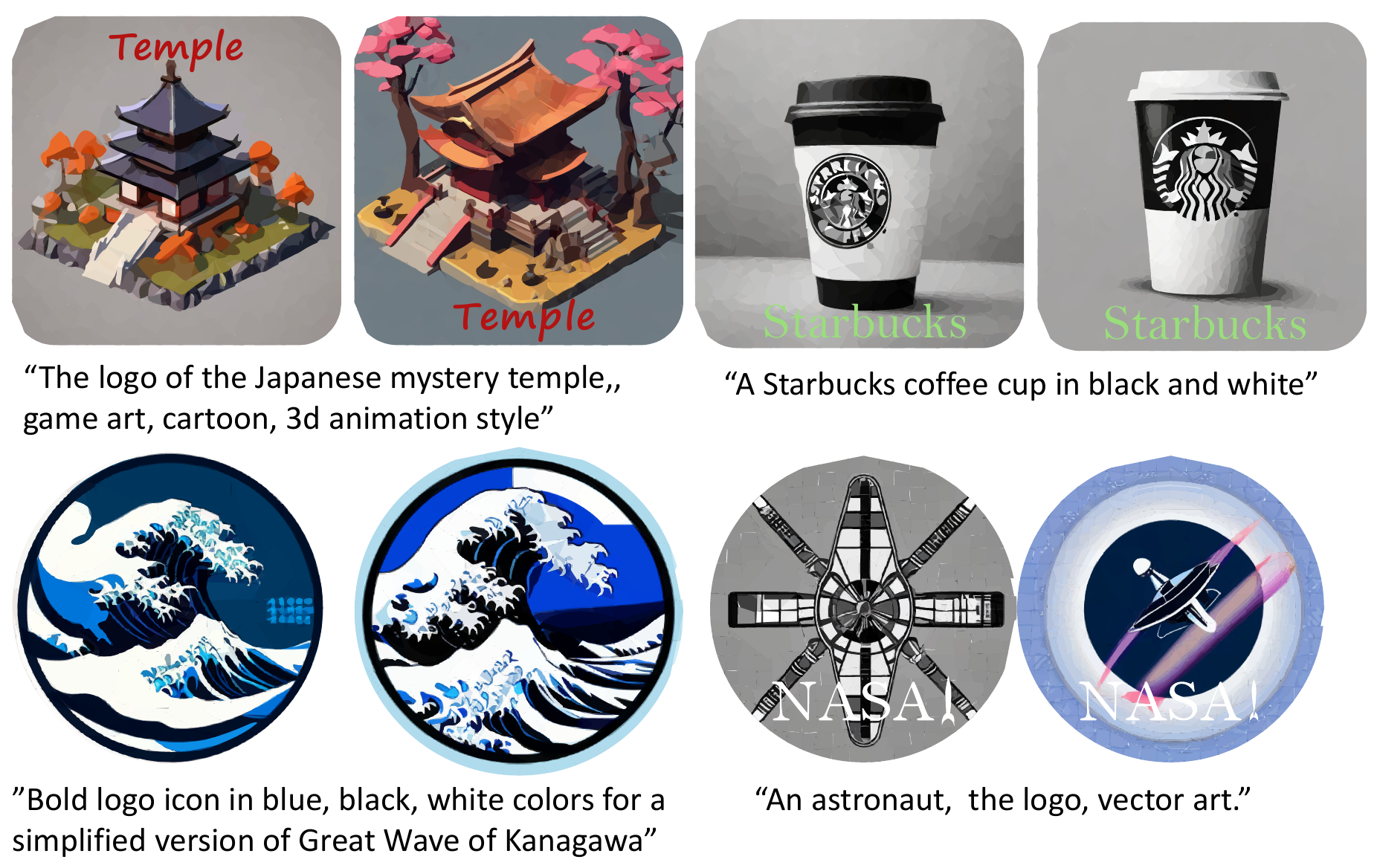}
\caption{
\textbf{Examples of icons created with SVGDreamer.}
SVGDreamer generates the main graphical content, which can then be combined with manually designed layouts and glyphs to form complete vector icons. The glyphs are manually added.
}
\label{fig:supp_icons}
\end{figure}

In addition to posters, SVGDreamer can also be used for icon design, as shown in Fig.~\ref{fig:supp_icons}. The examples include several common icon formats, such as badge-style emblems, circular logos, and product-branding compositions. In each case, SVGDreamer is used to generate the main graphical content, while the final icon layout is completed by combining the generated SVG assets with manually designed geometric containers and vectorized text.

A key observation from Fig.~\ref{fig:supp_icons} is that the generated graphic elements remain compatible with standard icon design structures. For instance, scene-centric content such as the temple examples can be embedded into emblem-style layouts, while object-centric results such as the astronaut and cup can be naturally composed with circular borders, product labels, or brand text. This shows that the outputs of SVGDreamer are not merely standalone SVG illustrations, but reusable vector assets that can be integrated into practical design workflows.

To construct the final icons, we first generate the core graphic content with SVGDreamer, then define polygonal or circular layouts using \texttt{def} tags in the SVG file, and finally append vectorized text paths to obtain complete vector icons. This demonstrates that SVGDreamer can serve not only as a text-to-SVG generator, but also as a practical asset creation tool for downstream vector design pipelines. The generated elements can be aligned with manually designed layouts and typography while preserving the editability, scalability, and compositional flexibility of vector representation.

%-----------------------------------Figure--------------------------------------
%-------------------------------------------------------------------------

\section{Implementation Details}
\label{sec:supp_implement}

Our method is built on a pretrained Stable Diffusion model~\cite{ldm_Rombach_2022}. We use the Adam optimizer with $\beta_1=0.9$, $\beta_2=0.9$, and $\epsilon=10^{-6}$ to optimize the SVG path parameters $\theta=\{P_i,C_i\}_{i=1}^{n}$. We adopt a learning-rate warm-up strategy. During the first 50 iterations, the learning rate for control points is gradually increased from $0.01$ to $0.9$, after which it decays exponentially from $0.8$ to $0.4$ over the remaining 650 iterations, for a total of 700 iterations. The learning rate for color parameters is set to $0.1$, and the learning rate for stroke width is set to $0.01$.

For the LoRA~\cite{lora_hu_2022} parameters, we use the AdamW optimizer with $\beta_1=0.9$, $\beta_2=0.999$, $\epsilon=10^{-10}$, and $lr=10^{-5}$. In most experiments, the number of particles is set to $k=6$, meaning that six particles participate simultaneously in VPSD (Section~\ref{sec:SVGDreamer}), LoRA updates, and ReFL updates. To balance diversity, prompt fidelity, and visual detail, we set the classifier-free guidance (CFG~\cite{classifierfree_2022_ho}) scale to $7.5$. During optimization, SVGDreamer requires at least 31\,GB of GPU memory on an NVIDIA V100 GPU to generate six SVGs in parallel.
Empirically, this setting provides a good balance between prompt fidelity and diversity. Larger CFG values tend to reduce diversity, while smaller values may destabilize optimization, especially in the early stage of vector parameter refinement.

For flat iconographic vector synthesis, we optimize both path control points and fill colors. During optimization, many paths either shrink to a very small area or converge to low opacity, making them effectively unused. To encourage path utilization and thus improve diversity and detail, we periodically reinitialize paths whose fill opacity or area falls below a threshold, following the strategy of VectorFusion~\cite{vectorfusion_jain_2023}. These discarded paths are removed from the current optimization state and reintroduced as randomly initialized colored circles placed on top of existing paths.

\section{Object Identification in SIVE Prompts}
\label{sec:supp_prompt_sive}

\begin{figure}[h]
\centering
\includegraphics[width=1.0\linewidth]{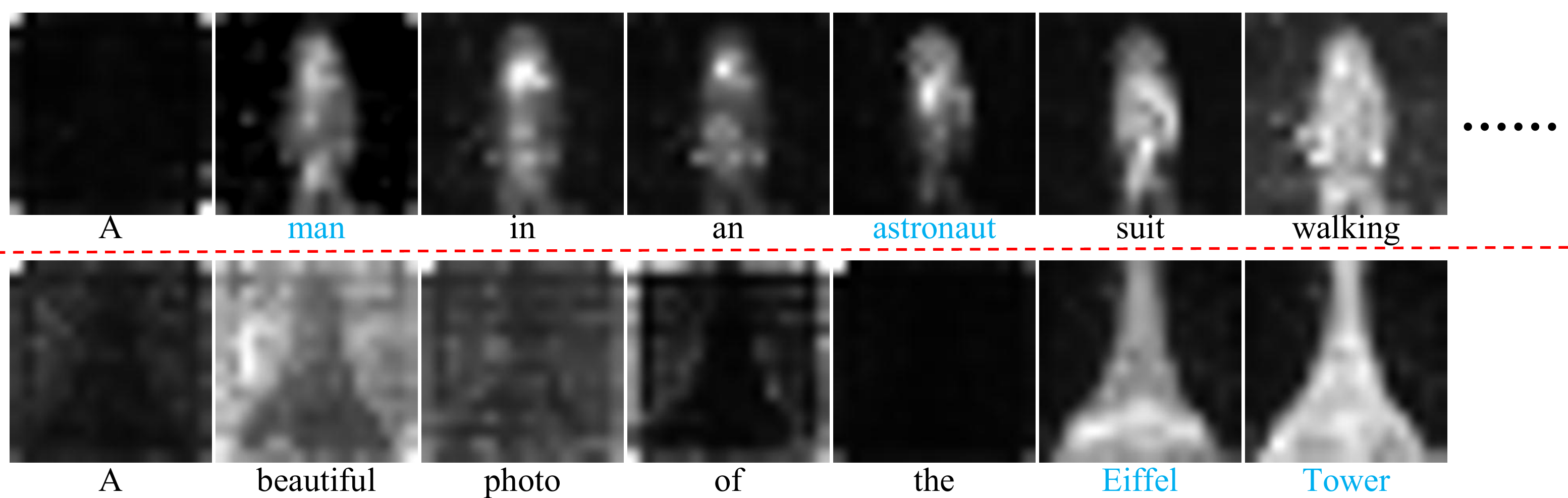}
\caption{
\textbf{Visualizations of the LDM cross-attention maps.}
Semantically related tokens often produce highly similar spatial responses, while multi-token object names may correspond to the same semantic region.
}
\label{fig:visual_attn}
\end{figure}

A practical issue in SIVE is that semantic objects do not always correspond one-to-one with individual tokens in the prompt. As shown in Fig.~\ref{fig:visual_attn}, multiple nouns within a sentence may refer to the same object, and a single object may also be described by multiple tokens. For example, in the first row, the tokens ``man'' and ``astronaut'' produce highly similar cross-attention maps and both correspond to the same foreground object. In the second row, the object ``Eiffel Tower'' is described by two tokens, ``Eiffel'' and ``Tower'', whose attention responses concentrate on the same semantic region.

In our experiments, we therefore do not adopt a dedicated token-selection strategy. Instead, we rely on the attention responses themselves: when semantically related tokens produce highly overlapping cross-attention maps, selecting either token usually leads to comparable vectorization results. This is sufficient for the prompts used in our experiments, where the main semantic objects typically exhibit clear and concentrated attention responses.

For more precise control, users may directly inspect the cross-attention maps associated with the input prompt and choose the tokens whose spatial responses best match the desired objects. In this sense, SIVE supports a visually grounded way of identifying semantic objects, where object selection is guided by attention localization rather than by fixed linguistic rules alone.

\section{Additional Ablation Studies}
\label{sec:supp_ablation}

In this section, we present additional ablation studies to further analyze the behavior of SVGDreamer and validate the effectiveness of its key design choices. Specifically, we examine the effects of the classifier-free guidance (CFG) weight, Reward Feedback Learning (ReFL), the number of vector particles, and the number of paths. These experiments provide a more comprehensive understanding of how different components and hyperparameters affect generation diversity, optimization stability, efficiency, and visual quality.

%-----------------------------------Figure--------------------------------------
\begin{figure}[t]
\centering
\includegraphics[width=1.0\linewidth]{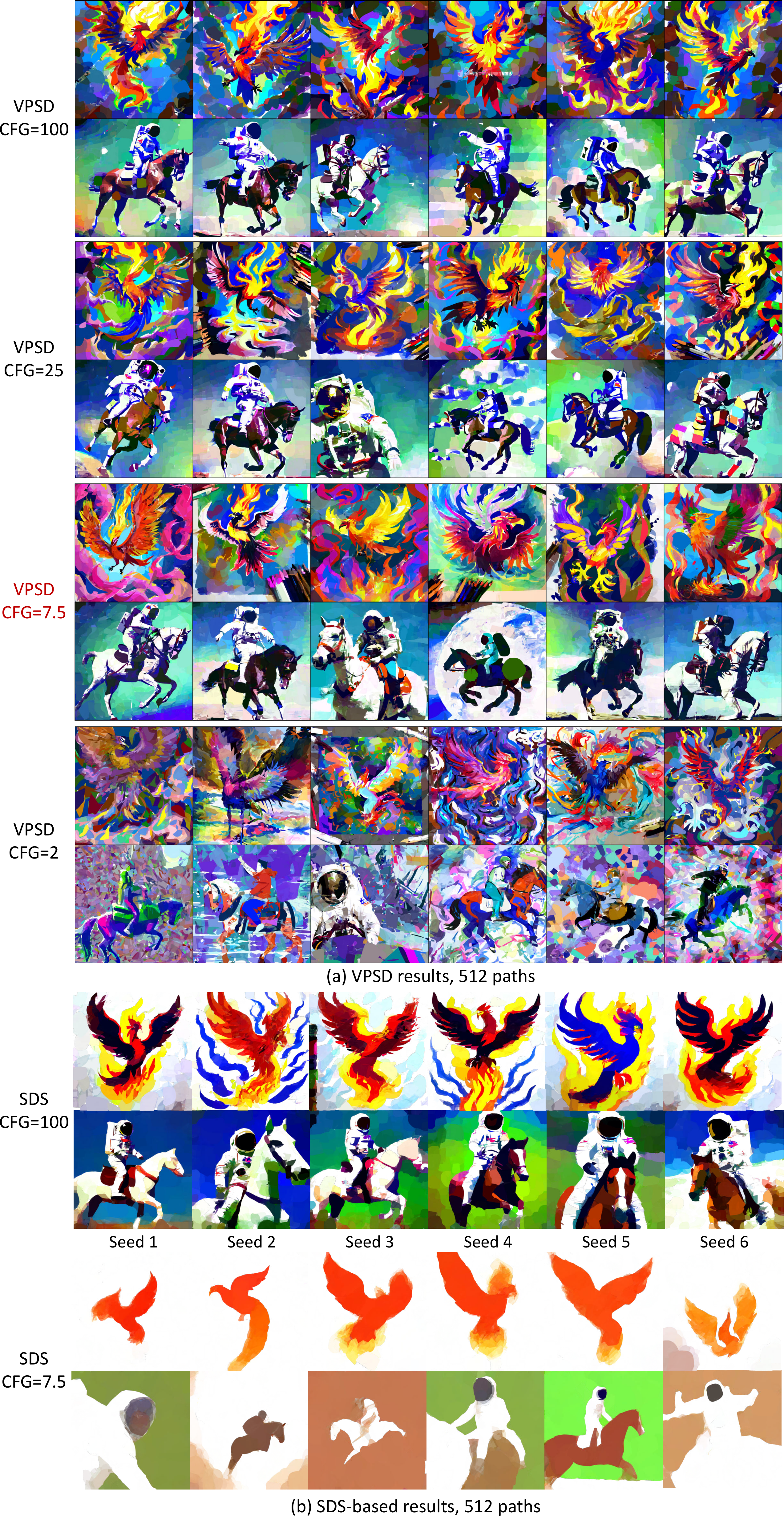}
\vspace{-1em}
\caption{
\textbf{Effect of different classifier-free guidance (CFG) weights on generation diversity.}
Smaller CFG values lead to higher diversity, but overly small CFG values reduce optimization stability. The prompt is ``A photograph of an astronaut riding a horse''.
} \label{fig:supp_cfg}
\vspace{-1em}
\end{figure}
%-------------------------------------------------------------------------
%-----------------------------------Figure--------------------------------------
\begin{figure*}[!t]
\centering
\includegraphics[width=0.85\linewidth]{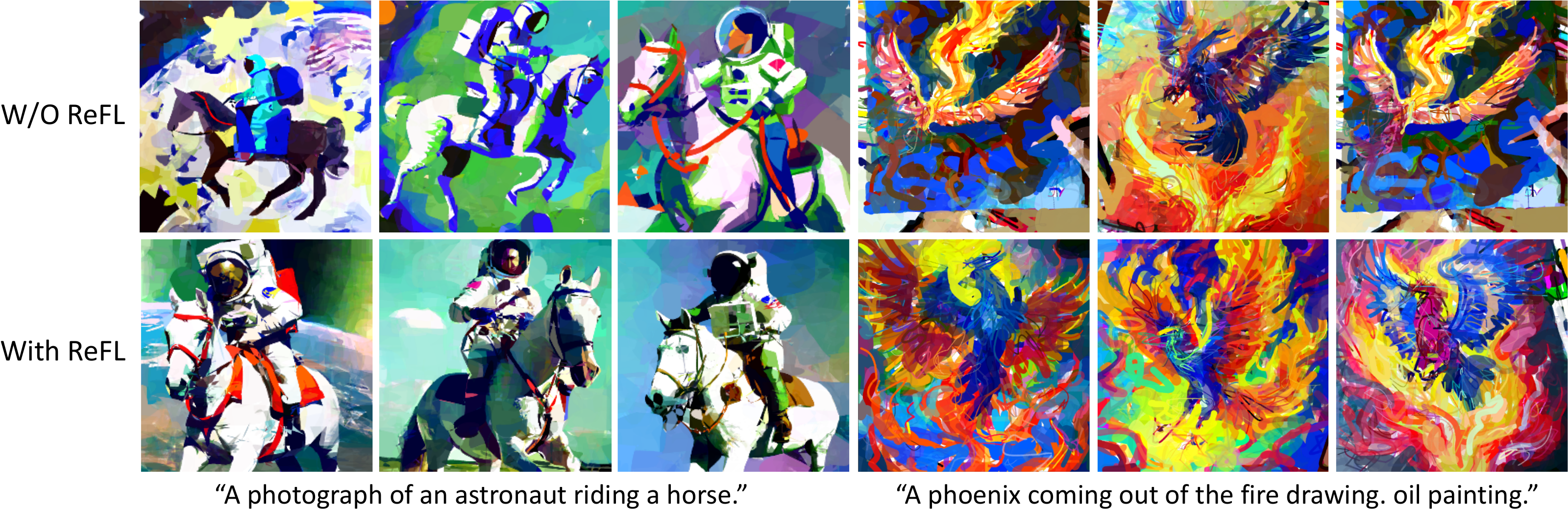}
\vspace{-0.5em}
\caption{
\textbf{Effect of Reward Feedback Learning (ReFL).}
Compared with the variant without ReFL, incorporating ReFL produces SVGs with improved visual quality, clearer semantic structure, and more coherent details across different prompts.
} \label{fig:supp_refl}
\vspace{-1em}
\end{figure*}
%-------------------------------------------------------------------------

\subsection{Ablation on CFG Weights}
\label{sec:supp_abl_CFG}

In this section, we investigate how the classifier-free guidance (CFG) weight~\cite{classifierfree_2022_ho} affects the diversity of the generated results. For VPSD, we use 6 particles and evaluate several CFG values. For LSDS~\cite{vectorfusion_jain_2023}, we perform four independent runs with different random seeds. The results are presented in Fig.~\ref{fig:supp_cfg}.

As shown in Fig.~\ref{fig:supp_cfg}, smaller CFG values generally lead to more diverse generations. We attribute this behavior to the fact that weaker guidance results in a broader effective distribution, allowing the optimization to explore a wider range of plausible modes. However, when the CFG weight becomes too small (e.g., CFG$=2$), the diffusion prior no longer provides sufficiently strong semantic guidance, which makes the optimization unstable and may produce unreasonable outputs. In our implementation, we therefore set CFG to $7.5$, which provides a practical balance between diversity and optimization stability.

It is also worth noting that SDS-based methods~\cite{vectorfusion_jain_2023,diffsketcher_xing_2023} degrade substantially under such low CFG values. By contrast, VPSD exhibits a more favorable trade-off between guidance strength and diversity, and can produce noticeably more diverse results simply by reducing the CFG weight.

This observation is consistent with the known role of CFG in diffusion-based generation: larger CFG weights improve prompt alignment but tend to concentrate the optimization on a small number of dominant modes, whereas smaller CFG weights encourage diversity at the cost of weaker semantic control. Compared with SDS-based methods, VPSD benefits more from this trade-off, likely because its particle-based formulation is better suited to exploring multiple plausible solutions.

\subsection{Ablation on ReFL}
\label{sec:supp_abl_refl}
% --------------Table--------------
\begin{table}[h]
\centering
\caption{\textbf{Efficiency of Reward Feedback Learning (ReFL) in SVGDreamer.} 
ReFL reduces the required optimization steps from $500$ to $300$ and nearly halves the runtime under different canvas sizes and path numbers.}
\label{tab:refl}
\resizebox{1.0\linewidth}{!}{
\begin{tabular}{c|c|c|c|c}
\toprule
Method & Canvas Size & Path Number & Iteration Steps & Time (min:sec) \\
\midrule
W/O ReFL & 224 $\times$ 224 & 128 & 500 & 13m15s \\
\midrule
W ReFL & 224 $\times$ 224 & 128 & 300 & 6m45s \\
\midrule
W/O ReFL & 600 $\times$ 600 & 256 & 500 & 14m21s \\
\midrule
W ReFL & 600 $\times$ 600 & 256 & 300 & 7m21s \\
\bottomrule
\end{tabular}
}
\end{table}
% ----------------------------

In ProlificDreamer~\cite{prolificdreamer_wang_2023}, only a subset of particles is used to update the LoRA network at each iteration. However, this strategy does not explicitly account for the current learning status of the LoRA network, which serves as an estimator of the variational distribution. In practice, this estimator may require many iterations to approach the target distribution, leading to slow convergence. In addition, the randomness introduced by particle initialization can cause suboptimal particles to dominate the early stage of training, which may adversely affect the final generation quality.

To alleviate this issue, we introduce Reward Feedback Learning (ReFL). Specifically, ReFL employs a pretrained reward model~\cite{imagereward_xu_2023} to score samples generated from the LoRA model, and then updates the LoRA model using reward-reweighted samples. This feedback mechanism encourages the estimator to place more emphasis on promising samples during optimization, rather than treating all intermediate samples equally.

As shown in Table~\ref{tab:refl}, ReFL reduces the required number of optimization steps from $500$ to $300$, leading to an almost $50\%$ reduction in runtime under both experimental settings. Meanwhile, Table~\ref{tab:quantitative} shows that ReFL further improves the aesthetic score of the generated SVGs. The qualitative improvements are also evident in Fig.~\ref{fig:supp_refl}, where ReFL produces results with better structure, clearer semantics, and more visually appealing details.

Overall, these results indicate that ReFL improves both efficiency and optimization robustness. By biasing the estimator toward higher-reward samples, ReFL suppresses the influence of low-quality intermediate results and helps the model converge to solutions that are both more visually appealing and more semantically faithful to the text prompt.

%-----------------------------------Figure--------------------------------------
\begin{figure*}[h]
\centering
\includegraphics[width=0.85\linewidth]{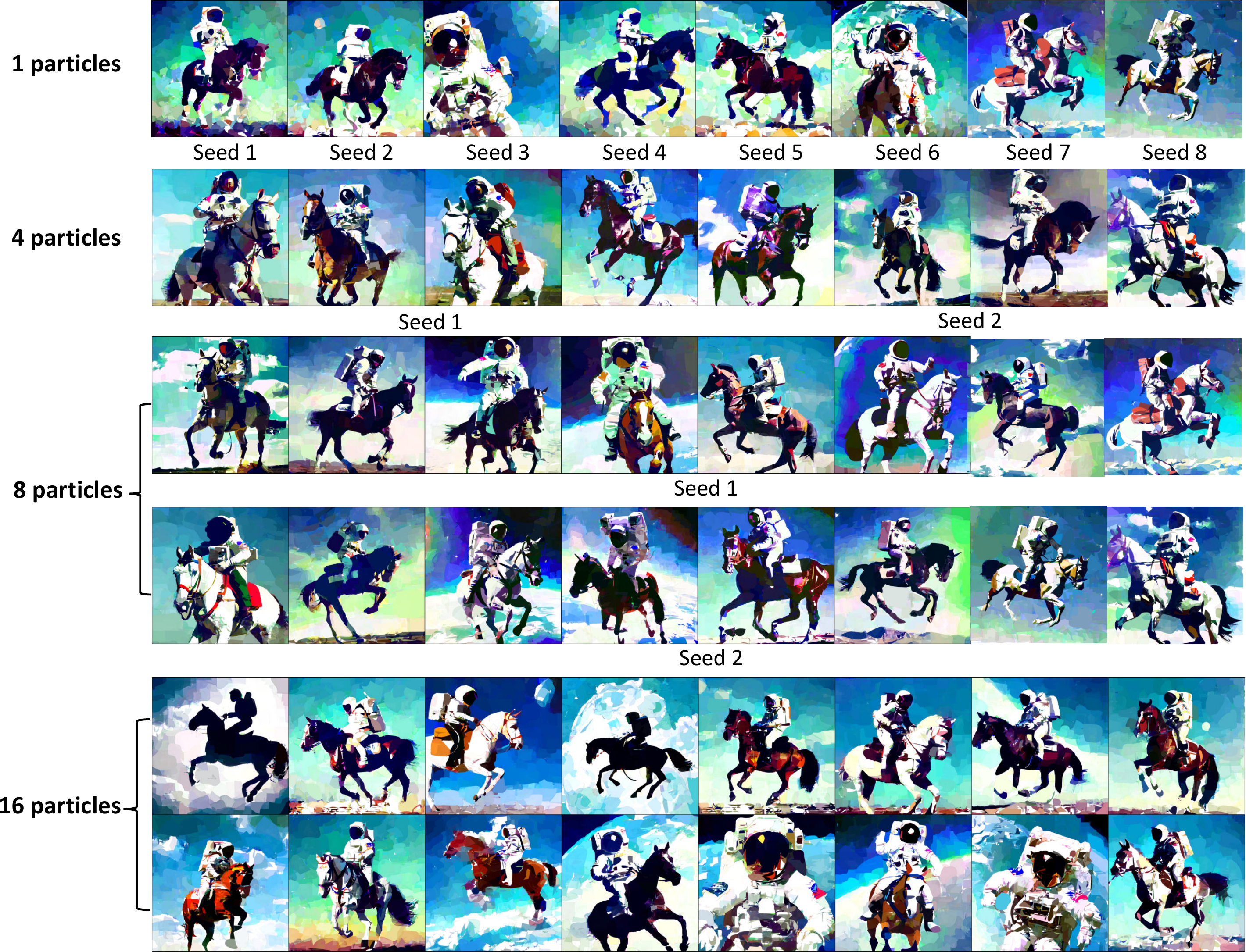}
\vspace{-0.5em}
\caption{\textbf{Effect of the number of vector particles.} 
Increasing the number of particles slightly improves generation diversity, while the overall visual quality remains relatively stable. The prompt is ``A photograph of an astronaut riding a horse''.}
\label{fig:supp_particles}
\vspace{-0.5em}
\end{figure*}
%-------------------------------------------------------------------------
\subsection{Ablation on the Number of Vector Particles}
\label{sec:supp_abl_n_particles}

We study the effect of the number of vector particles on generation performance. Specifically, we vary the number of particles among 1, 4, 8, and 16, and analyze how it influences the generated results. For all experiments, the CFG weight of VPSD is fixed to $7.5$.

As shown in Fig.~\ref{fig:supp_particles}, increasing the number of particles leads to slightly higher diversity in the generated results. Meanwhile, the overall visual quality remains relatively stable across different particle numbers, suggesting that VPSD is not overly sensitive to this hyperparameter in terms of generation quality.

Considering the substantial computational cost of optimizing vector primitive representations, as well as practical resource constraints, we use 6 particles in the main experiments as a compromise between diversity and efficiency. Although larger particle numbers can further improve diversity, the gain is relatively limited compared with the additional computational overhead.

%-----------------------------------Figure--------------------------------
\begin{figure}[h]
\centering
\includegraphics[width=1.0\linewidth]{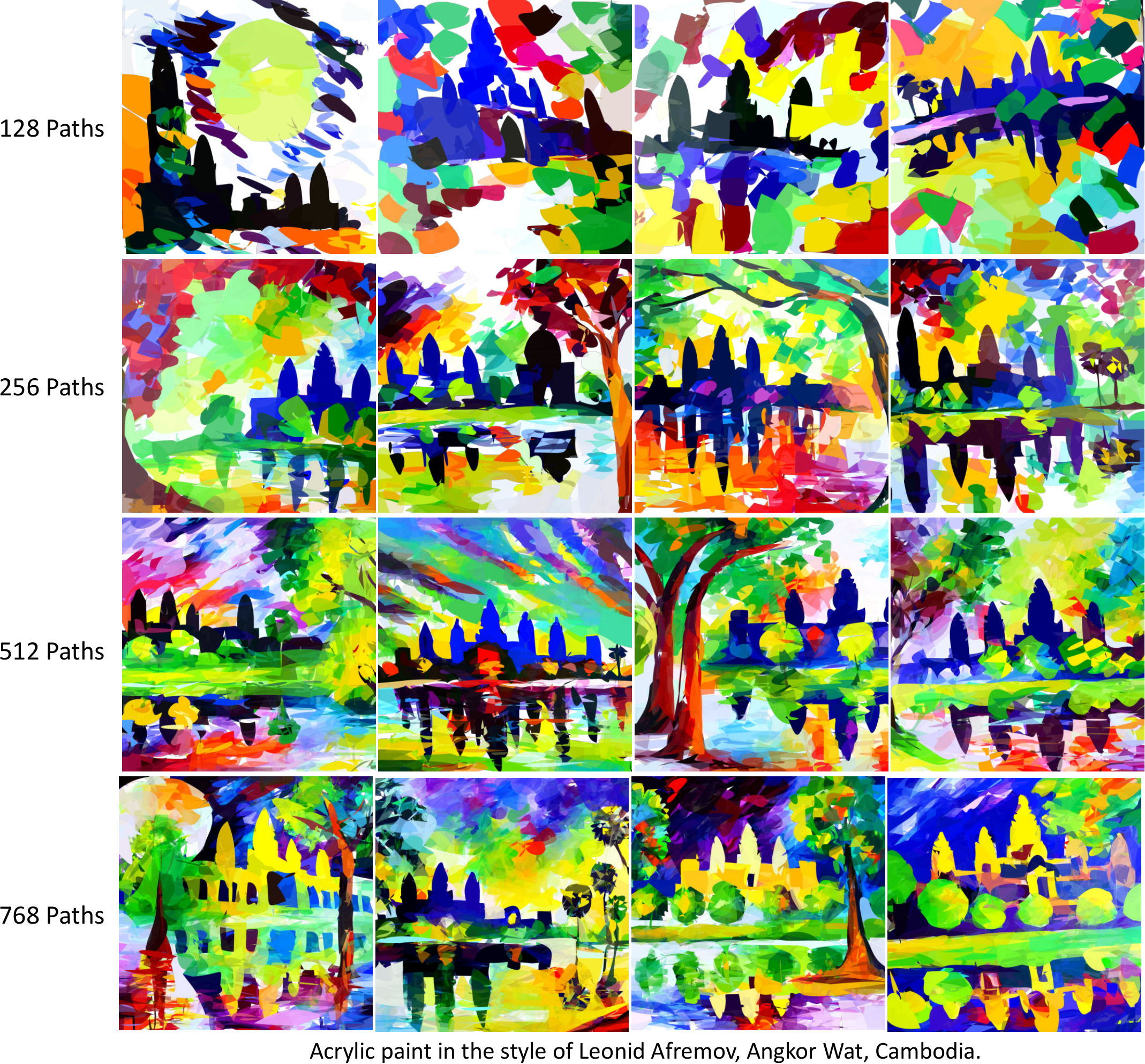}
\vspace{-1.5em}
\caption{
\textbf{Effect of the number of paths.}
Qualitative results with different path budgets. Increasing the number of vector paths improves structural completeness and local details, leading to SVGs that are more recognizable and visually coherent.
}
\label{fig:supp_num_paths}
\end{figure}
%-------------------------------------------------------------------------
\subsection{Ablation on the Number of Paths}
\label{sec:supp_abl_num_paths}

We analyze how the path budget affects SVG synthesis. Figure~\ref{fig:supp_num_paths} shows results generated with 128, 256, 512, and 768 paths under the same prompt and primitive setting.

A clear trend is that increasing the number of paths mainly improves the \emph{structural completeness} and \emph{local detail} of the generated SVGs. With 128 paths, the model captures only coarse color regions and a rough scene layout; object boundaries are weak, and many areas are represented by large abstract patches. With 256 paths, the global composition becomes more recognizable, and major semantic regions such as buildings, trees, and reflections begin to emerge. When the number of paths increases to 512 and 768, the scene becomes substantially more detailed: edges are sharper, the separation between foreground and background is clearer, and fine-grained elements such as reflections, foliage, and architectural contours are better preserved.

We also observe that a larger path budget generally leads to better semantic fidelity. With a small number of paths, the optimization tends to allocate most primitives to dominant color regions, leaving limited capacity for prompt-specific structures. As more paths become available, VPSD can represent both the global composition and local semantic details more effectively, resulting in outputs that better match the input prompt.

These results suggest that the number of paths is a key representational bottleneck in SVG synthesis. A limited path budget forces the model to trade detail for coverage, whereas a larger budget provides the geometric and chromatic flexibility needed for fine-grained vector generation.

\section{VPSD for 2D Image Synthesis}
\label{sec:supp_vpsd_2d_image}
%-----------------------------------Figure--------------------------------
\begin{figure}[h]
\centering
\includegraphics[width=1.0\linewidth]{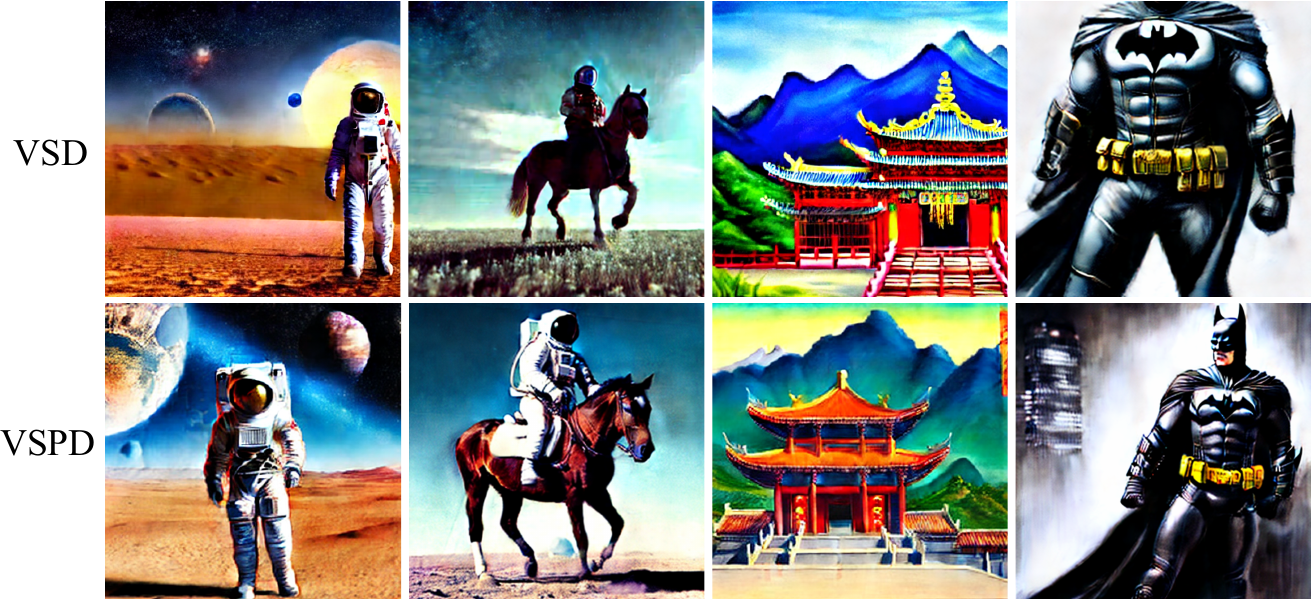}
\vspace{-1em}
\caption{\textbf{VPSD for 2D image synthesis.} Qualitative comparison between VSD and VPSD on 2D image synthesis. VPSD produces images with more complete structures, more stable compositions, and better visual quality.}
\label{fig:vpsd_2d}
\end{figure}
%-------------------------------------------------------------------------
Although VPSD is introduced in this work for text-to-SVG generation, it can also be extended to 2D image synthesis. This experiment is intended to demonstrate that the core idea of VPSD is not restricted to vector parameter spaces, but is also applicable to more general optimization-based visual generation tasks.

As shown in Fig.~\ref{fig:vpsd_2d}, VSD often produces results with incomplete object structures, unstable compositions, or inconsistent visual details. For example, some synthesized subjects appear partially missing, spatially misaligned, or visually less coherent. In contrast, VPSD generates images with more complete structures, more stable layouts, and better overall visual quality.

A possible reason for this improvement is that VPSD performs reward-guided particle refinement during optimization. By incorporating reward feedback, VPSD can better suppress low-quality intermediate solutions and steer the optimization toward samples that are both semantically plausible and aesthetically preferable. Although our primary focus is SVG synthesis, these results suggest that VPSD may generalize to other parameterized image generation problems where diversity, structural completeness, and perceptual quality are all important.

\begin{algorithm*}[h]
\caption{Vectorized Particle-based Score Distillation (VPSD)}
\label{algo:vpsd}
\begin{algorithmic}[1]
\Require Text prompt $y$; number of particles $k$ ($\geq 1$); number of SVG primitives $n$ ($\geq 1$); pretrained text-to-image diffusion model $\epsilon_{\phi}$; learning rate $\eta_p$ for SVG parameters; learning rate $\eta_e$ for LoRA parameters; pretrained reward model $r$~\cite{imagereward_xu_2023}; reward feedback strength $\lambda_r$.
\State \textbf{Initialize:} $k$ groups of SVG parameters $\{\theta^{(1)}, \cdots, \theta^{(k)}\}$, where each $\theta^{(i)}=\{(P_j^{(i)}, C_j^{(i)})\}_{j=1}^{n}$; a pretrained diffusion model $\epsilon_{\phi}$ with frozen parameters $\phi$; a LoRA~\cite{lora_hu_2022} estimator $\epsilon_{\phi_{\mathrm{est}}}$ parameterized by $\phi_{\mathrm{est}}$; and the pretrained reward model $r$.
\While{not converged}
    \State Randomly sample one particle $\theta \sim \{\theta^{(i)}\}_{i=1}^{k}$.
    \State Render the selected SVG parameters to obtain a raster image $x=\mathcal{R}(\theta)$.
    \State Update $\theta$ by
    \Statex \hspace{\algorithmicindent} $\theta \leftarrow \theta - \eta_p \,
    \mathbb{E}_{t,\epsilon,p,c}
    \left[
    \omega(t)
    \left(
    \epsilon_{\phi}(\mathbf{z}_t; y, t)
    -
    \epsilon_{\phi_{\mathrm{est}}}(\mathbf{z}_t; y, p, c, t)
    \right)
    \frac{\partial \mathbf{z}}{\partial \theta}
    \right]$
    \State Sample $w$ ($\leq k$) images using $\epsilon_{\phi_{\mathrm{est}}}(y)$.
    \State Update $\phi_{\mathrm{est}}$ by
    \Statex \hspace{\algorithmicindent} $\phi_{\mathrm{est}} \leftarrow \phi_{\mathrm{est}} - \eta_e \,
    \nabla_{\phi_{\mathrm{est}}}
    \left[
    \mathbb{E}_{\epsilon,t}
    \left\|
    \epsilon_{\phi_{\mathrm{est}}}(\mathbf{z}_t; y, p, c, t) - \epsilon
    \right\|_2^2
    +
    \lambda_r
    \mathbb{E}_{y,w}
    \left[
    \psi\!\left(r\!\left(y, g_{\phi_{\mathrm{est}}}(y)\right)\right)
    \right]
    \right]$
\EndWhile
\State \Return $\{\theta^{(1)}, \cdots, \theta^{(k)}\}$.
\end{algorithmic}
\end{algorithm*}

\begin{algorithm*}[h]
\caption{Semantic-driven Image Vectorization (SIVE) + VPSD}
\label{algo:sive_vpsd}
\begin{algorithmic}[1]
\Require Text prompt $y$; number of particles $k$ ($\geq 1$); number of SVG primitives $n$ ($\geq 1$); pretrained text-to-image diffusion model $\epsilon_{\phi}$; learning rate $\eta_p$ for SVG parameters; learning rate $\eta_e$ for LoRA parameters; pretrained reward model $r$~\cite{imagereward_xu_2023}; reward feedback strength $\lambda_r$.
\State \textbf{Initialize:} $k$ groups of SVG parameters $\{\theta^{(1)}, \cdots, \theta^{(k)}\}$, where each $\theta^{(i)}=\{(P_j^{(i)}, C_j^{(i)})\}_{j=1}^{n}$; and a pretrained diffusion model $\epsilon_{\phi}$.
\State Generate an initial sample from $\epsilon_{\phi}(y)$.
\State Extract the cross-attention map for the $i$-th text token:
\Statex \hspace{\algorithmicindent} $\mathcal{M}_{\mathrm{FG}}^i = \mathrm{softmax}\!\left(\frac{QK_i^{\top}}{\sqrt{d}}\right)$
\State Construct the background attention map:
\Statex \hspace{\algorithmicindent} $\mathcal{M}_{\mathrm{BG}} = 1 - \sum_{i=1}^{O} \mathcal{M}_{\mathrm{FG}}^i$
\State Construct the foreground and background masks:
\Statex \hspace{\algorithmicindent} $\hat{\mathcal{M}} = \left\{\{\hat{\mathcal{M}}_{\mathrm{FG}}^{o}\}_{o=1}^{O}, \hat{\mathcal{M}}_{\mathrm{BG}}\right\}$
\While{SIVE not converged}
    \State Update the initialized SVG parameters using the SIVE objective:
    \Statex \hspace{\algorithmicindent} $\theta^{(1)} \leftarrow \theta^{(1)} - \eta_p \nabla_{\theta}
    \sum_{i}
    \left(
    \hat{\mathcal{M}}_i \odot I - \hat{\mathcal{M}}_i \odot \mathbf{x}
    \right)^2$
\EndWhile
\State \textbf{Initialize:} a LoRA estimator $\epsilon_{\phi_{\mathrm{est}}}$ parameterized by $\phi_{\mathrm{est}}$, and the pretrained reward model $r$.
\While{VPSD not converged}
    \State Randomly sample one particle $\theta \sim \{\theta^{(i)}\}_{i=1}^{k}$.
    \State Render the selected SVG parameters to obtain a raster image $x=\mathcal{R}(\theta)$.
    \State Update $\theta$ by
    \Statex \hspace{\algorithmicindent} $\theta \leftarrow \theta - \eta_p \,
    \mathbb{E}_{t,\epsilon,p,c}
    \left[
    \omega(t)
    \left(
    \epsilon_{\phi}(\mathbf{z}_t; y, t)
    -
    \epsilon_{\phi_{\mathrm{est}}}(\mathbf{z}_t; y, p, c, t)
    \right)
    \frac{\partial \mathbf{z}}{\partial \theta}
    \right]$
    \State Sample $w$ ($\leq k$) images using $\epsilon_{\phi_{\mathrm{est}}}(y)$.
    \State Update $\phi_{\mathrm{est}}$ by
    \Statex \hspace{\algorithmicindent} $\phi_{\mathrm{est}} \leftarrow \phi_{\mathrm{est}} - \eta_e \,
    \nabla_{\phi_{\mathrm{est}}}
    \left[
    \mathbb{E}_{\epsilon,t}
    \left\|
    \epsilon_{\phi_{\mathrm{est}}}(\mathbf{z}_t; y, p, c, t) - \epsilon
    \right\|_2^2
    +
    \lambda_r
    \mathbb{E}_{y,w}
    \left[
    \psi\!\left(r\!\left(y, g_{\phi_{\mathrm{est}}}(y)\right)\right)
    \right]
    \right]$
\EndWhile
\State \Return $\{\theta^{(1)}, \cdots, \theta^{(k)}\}$.
\end{algorithmic}
\end{algorithm*}

\section{Algorithm for VPSD}
\label{sec:supp_algo_vpsd}

We summarize the algorithm of Vectorized Particle-based Score Distillation (VPSD) in \cref{algo:vpsd}. 
VPSD first initializes $k$ ($\geq 1$) groups of SVG parameters, a pretrained diffusion model $\epsilon_{\phi}$ parameterized by $\phi$, a LoRA-based estimator $\epsilon_{\phi_{\mathrm{est}}}$ parameterized by $\phi_{\mathrm{est}}$, and a pretrained reward model $r$.
Note that the pretrained diffusion model remains frozen, while only the LoRA parameters are updated during optimization~\cite{lora_hu_2022}.

At each iteration, VPSD randomly selects one SVG parameter set $\theta$ from the particle set and renders it into a raster image $x$. The selected parameter set $\theta$ is then updated using the VPSD objective. Meanwhile, samples generated by $\epsilon_{\phi_{\mathrm{est}}}(y)$ are used to update the estimator parameters $\phi_{\mathrm{est}}$. This process is repeated until convergence, and the algorithm finally returns $k$ groups of SVG parameters as the output.

\cref{algo:sive_vpsd} summarizes the full pipeline that combines VPSD with SIVE (Semantic-driven Image Vectorization). This algorithm shares the same initialization as VPSD, but it additionally requires an initial sample generated by the diffusion model $\epsilon_{\phi}$ from the text prompt $y$. During the sampling process, the corresponding cross-attention maps are extracted. Based on these attention maps, the algorithm constructs foreground and background masks, optimizes the SVG parameters with the SIVE objective, and then further refines them using VPSD.

\end{document}